\documentclass[final,3p,times,twocolumn]{elsarticle}

\usepackage{amssymb}
\usepackage{amsmath,amsfonts}
\usepackage{algorithmic}
\usepackage{algorithm}
\usepackage{array}
\usepackage[caption=false,font=normalsize,labelfont=sf,textfont=sf]{subfig}
\usepackage{textcomp}
\usepackage{stfloats}
\usepackage{url}
\usepackage{verbatim}
\usepackage{graphicx}
\usepackage{color}
\usepackage{xcolor}
\usepackage{booktabs}
\usepackage{multirow}
\usepackage{subfig}
\usepackage{pifont}
\usepackage{makecell}
\usepackage{hyperref}
\usepackage{color}
\usepackage{subcaption} 

\usepackage{lineno}

\journal{Applied Soft Computing}

\begin{document}
\begin{frontmatter}


 \title{TAS-TsC: A Data-Driven Framework for Estimating Time of Arrival Using Temporal-Attribute-Spatial Tri-space Coordination of Truck Trajectories}


\author[1]{Mengran~Li}
\ead{limr39@mail2.sysu.edu.cn}

\author[1]{Junzhou~Chen}%
\ead{chenjunzhou@mail.sysu.edu.cn}

\author[2]{Guanying~Jiang}
\ead{guanyingjiang@gmail.com}

\author[3]{Fuliang~Li}
\ead{tjfulianglee@gmail.com}

\author[1]{Ronghui~Zhang\corref{cor1}%
}
\ead{zhangrh25@mail.sysu.edu.cn}

\author[4]{Siyuan~Gong}
\ead{sgong@chd.edu.cn}

\author[5]{Zhihan~Lv}
\ead{lvzhihan@gmail.com}

\cortext[cor1]{Corresponding author.}
\tnotetext[t1]{Our manuscript was first submitted to Applied Soft Computing on April 22, 2024.}

\tnotetext[t1]{This project is jointly supported by National Natural Science Foundation of China (Nos.52172350, 51775565), Guangdong Basic and Applied Research Foundation (Nos.2021B1515120032, 2022B1515120072), Guangzhou Science and Technology Plan Project (Nos.2024B01W0079, 202206030005), Nansha Key RD Program (No.2022ZD014), Science and Technology Planning Project of Guangdong Province (No.2023B1212060029).}

\affiliation[1]{organization={Guangdong Key Laboratory of Intelligent Transportation System, School of Intelligent Systems Engineering, Sun Yat-sen University},
                city={Guangzhou},
                postcode={510275}, 
                state={Guangdong},
                country={China}}

\affiliation[2]{organization={Baidu Inc},
                postcode={100085}, 
                city={Beijing},
                country={China}}

\affiliation[3]{organization={Department of Electrical Engineering, The Hong Kong Polytechnic University},
                country={ Hong Kong Special Administrative Region of China}}

\affiliation[4]{organization={School of Information and Engineering, Chang’an University},
                city={Xi'an},
                postcode={710064}, 
                state={Shanxi}, 
                country={China}}

\affiliation[5]{organization={Department of Game Design, Faculty of Arts, Uppsala University},
                city={Uppsala},
                postcode={75236}, 
                country={Sweden}}

\begin{abstract}
{
Accurately estimating time of arrival (ETA) for trucks is crucial for optimizing transportation efficiency in logistics. GPS trajectory data offers valuable information for ETA, but challenges arise due to temporal sparsity, variable sequence lengths, and the interdependencies among multiple trucks. To address these issues, we propose the Temporal-Attribute-Spatial Tri-space Coordination (TAS-TsC) framework, which leverages three feature spaces—temporal, attribute, and spatial—to enhance ETA. Our framework consists of a Temporal Learning Module (TLM) using state space models to capture temporal dependencies, an Attribute Extraction Module (AEM) that transforms sequential features into structured attribute embeddings, and a Spatial Fusion Module (SFM) that models the interactions among multiple trajectories using graph representation learning.} These modules collaboratively learn trajectory embeddings, which are then used by a Downstream Prediction Module (DPM) to estimate arrival times. We validate TAS-TsC on real truck trajectory datasets collected from Shenzhen, China, demonstrating its superior performance compared to existing methods.
\end{abstract}



\begin{keyword}
Estimating Time of Arrival ,
GPS Trajectory Data,
Temporal Learning,
Attribute Extraction,
Spatial Fusion,
Tir-space Coordination
\end{keyword}

\end{frontmatter}

\section{Introduction}

In global trade, accurately estimating time of arrival (ETA) is crucial for optimizing warehouse management, production scheduling, and maintaining balance between supply and demand in the logistics and transportation industry \cite{hesse2004transport, nadi2021short}. Over the last decade, ETA has transitioned from traditional statistical approaches to advanced data-driven models, harnessing the power of machine learning and deep learning technologies. These models utilize historical trajectory data to predict arrival times with greater precision \cite{gutierrez2021data, zhang2011data, li2023sequence}. 

\begin{figure*}[h]
\centering
\includegraphics[width=0.65\textwidth]{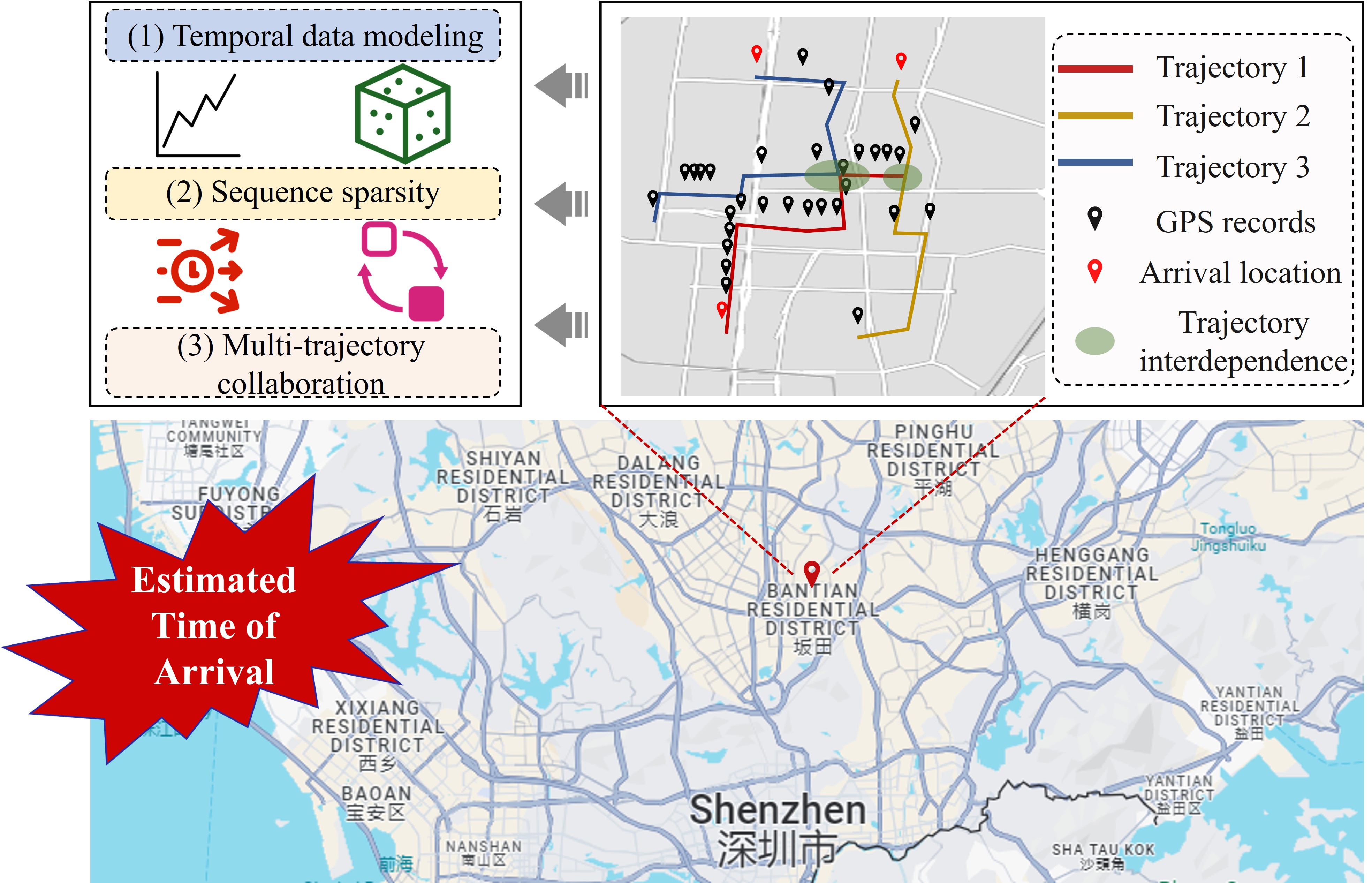}
\caption{Illustration of coordination of truck GPS trajectories in complex logistics scenarios.}
\label{fig0}
\end{figure*}

Current methodologies extract features from various data sources, including map-matched road segments \cite{huang2021bus, shi2021prediction, sun2020codriver, han2021multi}, origin-destination (OD) information \cite{zhou2023inductive, zhang2023delivery, zhang2024estimating, zhang2024package}, and GPS coordinates \cite{wang2018will, lin2019path, wang2022fine}. Map-matching techniques offer explicit details like road names and traffic conditions, while OD data provides contextual information, such as sender and recipient addresses, frequently used in e-commerce ETA \cite{de2021end, li2021unsupervised}. Among these, GPS data stands out for its ability to deliver real-time insights into vehicle movements, including position, speed, and direction, making it highly adaptable to dynamic conditions. With the advancement of sensor technologies and Intelligent Transportation Systems (ITS), GPS data enables valuable insights into large-scale traffic patterns and driving behavior. This continuous, real-time tracking captures implicit details of environmental interactions, which can be leveraged by advanced deep learning models to extract richer features, without relying on explicit map or traffic information \cite{wang2017computing}. Despite these advances, significant challenges remain in leveraging GPS data for accurate ETA:

\textbf{(1) Effective temporal trajectory data modeling.} Temporal model requires algorithms to handle large amounts of sequence data, identify time-dependent features such as dwell time, speed, and direction, as well as understand how these features change over time. Thus, effectively modeling these spatiotemporal features to predict road travel time remains a highly challenging task. Fortunately, the development of deep learning technologies over the past decade has led to the evolution of various neural network architectures, especially Convolutional Neural Networks (CNN) \cite{zhao2017convolutional, jin2022multivariate}, Recurrent Neural Networks (RNN) \cite{connor1994recurrent, borovykh2017conditional}, LSTM \cite{li2019ea, siami2019performance}, and Transformer \cite{qin2017dual, zhou2021informer}. These architectures have shown significant advantages in modeling real-world temporal sequence data. However, in complex time series analysis scenarios, methods like Transformers struggle with model computation on large-scale data due to their quadratic complexity.

\textbf{(2) Data sparsity and variable sequence lengths.} GPS signal instability, device malfunctions, and inconsistent data collection intervals lead to sparse trajectory data within the same range, with varying sequence lengths for different trajectories. As illustrated in Figure \ref{fig0}, even under the same spatiotemporal conditions, the GPS data recorded by different trajectories show irregularities. For instance, compared to the relatively regular Trajectory 1, Trajectory 2 lacks records during certain periods, and the recording intervals of Trajectory 3 are extremely sparse. This irregular data presents challenges in mapping trajectories to a learning space with overall structure and consistency \cite{liu2012mining, xing2022traffic}. To address this issue, existing research has attempted improvements. For example, Tang et al. \cite{tang2018citywide} and Huang et al. \cite{huang2022context} proposed a tensor-based urban spatiotemporal model aimed at uncovering the underlying structure of sparse tensors and estimating missing information. This method enhances the accuracy and comprehensiveness of ETA by simulating the travel times of different road segments under various traffic conditions and their corresponding occurrence probabilities. However, methods like tensor decomposition, while useful, increase computational complexity and may still not fully address the variable sequence lengths or capture the true dynamics of the data.

\textbf{(3) The collaborative operation of multiple trajectories.} In real-world logistics scenarios, the interaction and collaborative operation of multiple truck trajectories significantly affect the estimation of arrival times. {As shown in Figure \ref{fig0}, when trucks encounter each other at intersections, such as at a crossroad or during merging lanes, their trajectories are not independent but interdependent, influencing each other's travel times \cite{jin2023survey, wang2020deep}.} This complex interrelation and heterogeneity among trajectories pose significant challenges for modeling and arrival time estimation \cite{jin2023spatio}. It is crucial not only to identify and process the independent behavior of each truck but also to consider their dynamic interactions, which is a key factor in logistics optimization and planning. Notably, Graph Representation Learning (GRL) has emerged in recent years for learning representations of non-Euclidean data \cite{wu2020comprehensive, liu2022graph, jin2023expressive}. GRL excels at capturing a wide range of complex relationships, including the connections between attribute variables and sequence dependencies \cite{jin2023survey}, thus aiding in exploring real-world trajectory interactions. Furthermore, Ma et al. \cite{ma2022multi} and Qiu et al. \cite{qiu2023internet} used Graph Neural Networks (GNNs) \cite{kipf2016semi, velivckovic2017graph, hamilton2017inductive} to model sparse road networks and trajectories, demonstrating good performance. However, these methods must match the road network and may not effectively capture less obvious but crucial spatial and attribute interactions.

Given these challenges, there is a clear need for an integrated approach that can effectively capture temporal dynamics, handle data sparsity, and model interactions among multiple trajectories. To address these issues, we introduce the Temporal-Attribute-Spatial Tri-space Coordination (TAS-TsC) framework. The TAS-TsC framework integrates temporal, attribute, and spatial dimensions to address these challenges holistically. It consists of three modules: the Temporal Learning Module (TLM), the Attribute Extraction Module (AEM), and the Spatial Fusion Module (SFM), each designed to tackle specific aspects of the ETA problem.

The TLM tackles temporal data modeling challenges by adopting a State Space Model (SSM), specifically Mamba \cite{gu2023mamba}. Mamba captures long-term dependencies and dynamic changes in time series data, improving our understanding of how factors such as time, location, speed, and direction impact arrival times. By focusing on contextual information, Mamba effectively addresses data gaps or inconsistencies, and its linear complexity enhances model training performance compared to traditional architectures like Transformers.

To handle data sparsity and variable sequence lengths, traditional methods often rely on zero-padding or mean-padding, which may overlook critical trends. In contrast, TLM leverages Mamba to capture long-term dependencies and update hidden states at each time step. Meanwhile, at the feature level, AEM employs statistical methods to extract trajectory attributes, revealing the global feature distribution. AEM transforms sparse and variable-length data into unified representations that accurately reflect the attributes of the trajectories.

To account for the collaborative influence of multiple trajectories, SFM constructs a spatiotemporal relation graph to represent interactions and spatial relationships among truck trajectories. Using a graph diffusion approach, it integrates attribute and temporal features to capture spatial embeddings that represent high-order structural information within trajectories.

For model training, we implement a tri-space coordinated self-supervised learning approach that combines embedding learning with graph structure optimization. This optimization method enhances the similarity between the original trajectories and temporal feature embeddings while regularizing the correlation matrix derived from the hybrid feature embeddings. Finally, our downstream prediction module employs residual connections \cite{he2016deep} and Histogram-based Gradient Boosting (HGB) \cite{ke2017lightgbm} to accurately estimate truck arrival times.

The main contributions of this paper are as follows:

\begin{itemize}
    \item Addressing the challenge of integrating diverse data dimensions, we propose TAS-TsC, an innovative framework that merges temporal, attribute, and spatial dimensions of GPS trajectory data to enhance ETA in logistics transportation.
    
    \item Tackling the issue of temporal sparsity in trajectory data, our TLM utilizes a state space model based on Mamba \cite{gu2023mamba} to deeply encode temporal embeddings and capture contextual information of trajectories, thereby improving the accuracy of arrival time estimations.
    
    \item To enhance the semantic richness of the features and consider the spatial structural information inherent among multiple trajectories, the AEM employs feature engineering to transform implicit trajectory features into structured attribute embeddings, while the SFM captures essential spatial embeddings using a graph feature diffusion.
    
    \item We demonstrate the practical effectiveness of our framework through rigorous testing on real-world datasets, showing that TAS-TsC significantly outperforms existing methods in ETA.
\end{itemize}

This paper is structured as follows: Section \ref{s2} reviews related research on arrival time estimation. Section \ref{s3} details our employed frameworks. Section \ref{s4} discusses experimental results, and Section \ref{s5} concludes the paper and explores future research avenues.

\section{Related Work}\label{s2}
We first review the progression of ETA methodologies, highlighting key developments and limitations. We then examine the evolution of time-series learning methods, focusing on how these approaches contribute to the predictive accuracy and efficiency of ETA models.

\subsection{Estimating Time of Arrival}
Early ETA methods relied on traditional machine learning techniques, which provided foundational methods for improving prediction accuracy. For example, Guin \cite{guin2006travel} used seasonal ARIMA models for travel time prediction. In later years, methods like K-Nearest Neighbors (KNN) \cite{yu2011bus} and Support Vector Machines (SVM) \cite{cao2003support, cini2023taming} were adopted for their robustness in regression tasks. More advanced techniques like Random Forests \cite{yu2018prediction} and Gradient Boosting Models (GBM) \cite{zhang2015gradient, xia2017traffic} showed improved accuracy by capturing complex data relationships. These models, combined with Bayesian Optimization (BO), demonstrated strong predictive capabilities in applications like train delay prediction \cite{shi2021prediction, huang2022bus}.

As ETA research progressed, ensemble learning and graph-based methods became popular for enhancing predictive power. Ensemble learning models, such as the ones by Zhong et al. \cite{zhong2020bus}, employed multiple predictors to improve prediction accuracy. Meanwhile, graph-based models, leveraging Graph Neural Networks (GNNs) like Graph Convolutional Networks (GCNs) \cite{kipf2016semi}, GraphSAGE \cite{hamilton2017inductive}, and Graph Attention Networks (GATs) \cite{velivckovic2017graph}, became essential in ETA due to their ability to model complex spatiotemporal relationships. These models proved effective for urban transit systems by utilizing network connectivity information \cite{ma2022multi, qiu2023internet, li2024redundancy}.

Recent years have seen the application of deep learning to ETA with models like Convolutional LSTMs \cite{petersen2019multi} and hybrid trajectory networks \cite{lin2019path}, allowing for more granular analysis of trajectory data. These methods are particularly useful for multi-city and multi-regional ETA due to their ability to learn from diverse datasets \cite{wang2018will, wang2022fine}. In 2021, Han et al. \cite{han2021multi} introduced a multi-semantic path representation learning method that further enhanced the accuracy of travel time estimation by capturing fine-grained semantic information within trajectories.

Most recently, tensor decomposition methods, like those proposed by Huang et al. \cite{huang2022context}, and multi-faceted route representation learning by Liao et al. \cite{liao2024multi}, have enabled context-aware ETA that adapt to varying conditions by accounting for both spatial and temporal dependencies within trajectory data. These advancements underscore the importance of comprehensive representation learning to address the inherent variability in travel time data across different regions and traffic conditions.

Despite advancements, current ETA methods face limitations. Traditional models lack robustness in capturing complex temporal-spatial relationships, especially with sparse, variable-length data. While deep learning and graph-based models improve accuracy, they demand high computational resources.

\subsection{Time-Series Learning Methods}

Effective Estimating Time of Arrival (ETA) prediction relies heavily on robust time-series learning methods capable of handling complex temporal dependencies. Traditional statistical models like Autoregressive Integrated Moving Average (ARIMA) \cite{box1970distribution, zivot2006vector} and Vector Autoregressive (VAR) models \cite{biller2003modeling, rady2021time} are widely used for univariate and multivariate time series data. However, they face limitations in handling sparse, irregular, and variable-length data sequences commonly found in transportation systems \cite{guin2006travel}.

In recent years, deep learning has significantly advanced time-series learning. Recurrent Neural Networks (RNNs) and their variants like Long Short-Term Memory (LSTM) and Gated Recurrent Units (GRU) have traditionally been used to capture temporal dependencies in sequence data \cite{petersen2019multi, noman2021towards}. However, RNNs face critical limitations in ETA tasks, especially with long GPS trajectories. RNNs are prone to vanishing gradients, making it difficult to retain long-term information effectively. Additionally, their sequential processing nature limits parallelization, resulting in high computational costs on large-scale datasets, as is common in ETA applications. Transformers \cite{vaswani2017attention} partially address these issues by allowing parallel processing and capturing long-range dependencies through self-attention. However, the quadratic complexity of the attention mechanism makes Transformers computationally intensive and memory-demanding, especially for the long sequences typical of GPS data \cite{gu2023mamba}. This complexity hinders their scalability in ETA applications, where sequence lengths often vary widely.

Therefore, current methods have their limitations. For instance, Transformers that rely on attention mechanisms are limited in their application to long sequences due to their complexity; and the main challenge for graph-based learning methods lies in capturing information from dynamic graph structures where connectivity changes over time. For estimating arrival times, we require a method capable of handling both sparse, long time series and leveraging associated relationships.

\section{Methodology}\label{s3}

\begin{figure*}[t]
\centering
\includegraphics[width=1\textwidth]{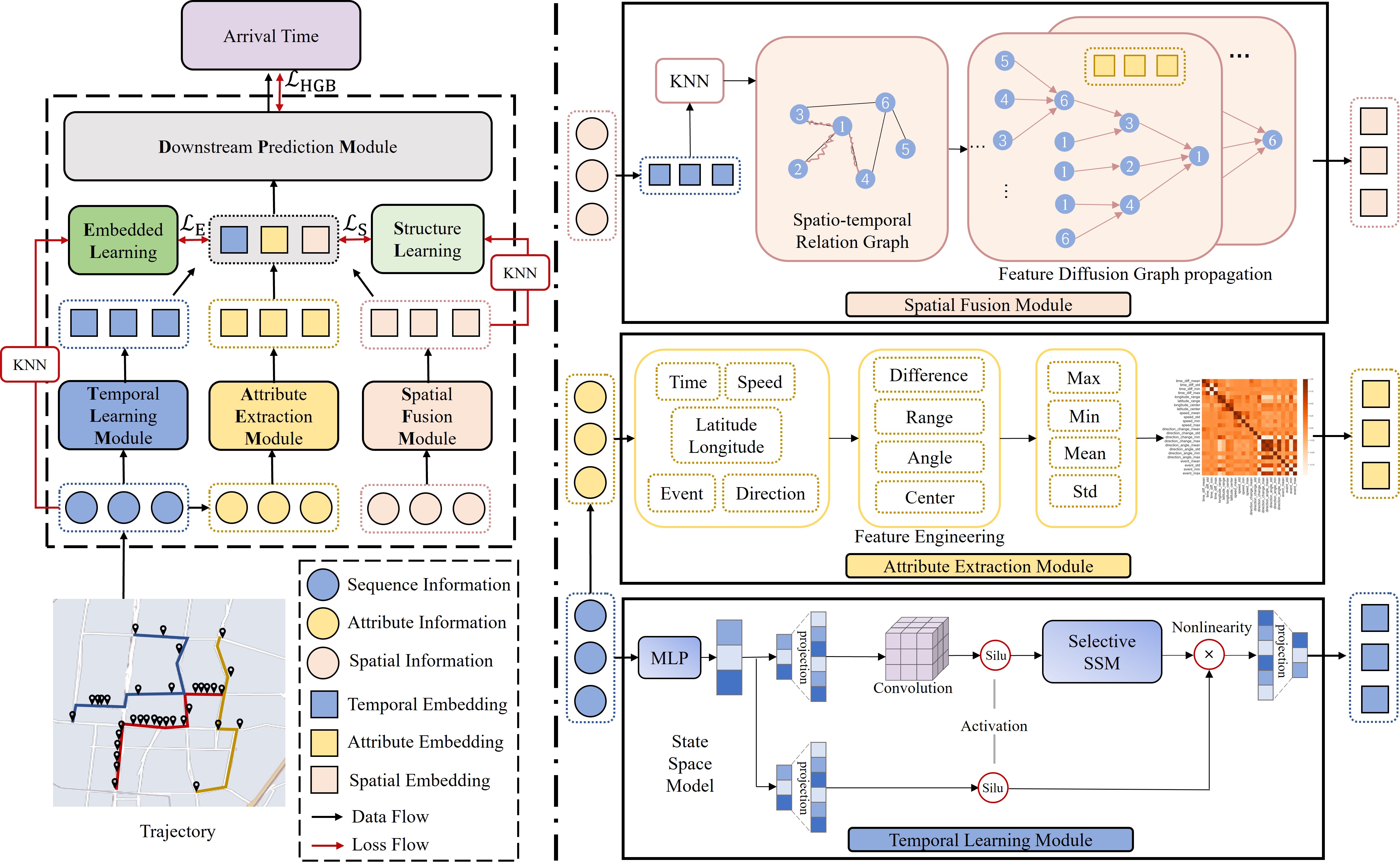}
\caption{{The overall structure of TAS-TsC framework. TLM captures temporal patterns in trajectory data using state-space modeling. AEM creates structured embeddings from trajectory attributes like speed and direction through feature engineering. SFM builds a spatiotemporal graph to model interactions among trajectories. DPM combines the outputs from TLM, AEM, and SFM using embedded and structure learning, and then uses HGB to predict truck arrival times.These modules work together to enhance ETA accuracy by leveraging temporal, attribute, and spatial data.}}
\label{fig1}
\end{figure*}

For ETA of trucks in logistics scenarios, we propose the Temporal-Attribute-Spatial Tri-space Coordination (TAS-TsC) framework, as depicted in Figure \ref{fig1}. Our framework encodes three sets of embeddings—temporal, attribute, and spatial—captured from the trajectory information through Temporal Learning Module (TLM), Attribute Extraction Module (AEM), and Spatial Fusion Module (SFM), respectively. These are then fused for integrated learning, with the final ETA through the Downstream Prediction Module (DPM).

\subsection{Preliminaries}
To estimate the arrival time based on vehicle trajectory data, we define a collection of variable-length trajectories gathered over a specific period. Let this set of trajectories be denoted as \( Tr = \{Tr_1, Tr_2, \dots, Tr_N\} \), where \( N \) represents the total number of trajectories. Each trajectory \( Tr_n \) (for \( n \leq N \)) consists of an ordered set of temporal data points, formally represented as \( Tr_n = \{P_1, P_2, \dots, P_{M_n}\} \), where \( M_n \) indicates the length of the \( n^{th} \) trajectory, which can vary from one trajectory to another.

Each data point \( P_i \) within a trajectory encapsulates multiple attributes structured as a feature vector. {This vector includes the timestamp \( t_i \), geographic coordinates (latitude \( \phi_i \) and longitude \( \lambda_i \)), speed \( v_i \), direction \( \theta_i \), and special event indicator \( e_i \). Thus, the feature vector for each point is defined as \( [t_i, \lambda_i, \phi_i, v_i, \theta_i, e_i] \), giving a total of six features (\( F = 6 \)) for each data point.}

Each trajectory \( Tr_n \) is associated with an arrival time label \( Y_n \). The objective of this study is to estimate \( Y \) by mapping \( Tr \) to \(\hat{Y}\) through a function \( \mathcal{F} \) that integrates multiple modules to minimize the difference between \(\hat{Y}\) and the actual arrival time \( Y \).

\subsection{Temporal Learning Module with State Space Model}

In time series forecasting, models like Transformers \cite{vaswani2017attention} demonstrate strong capabilities in capturing temporal dependencies, but their computational complexity of \(O(n^2)\) limits their ability to model longer sequences. To address this, state space models (SSMs) \cite{gu2021efficiently, fu2022hungry} have become increasingly popular due to their approximately linear complexity, enabling efficient modeling of complex relationships over long sequences. Our Temporal Learning Module (TLM) leverages a selected state space model (Mamba) \cite{wang2023selective, gu2023mamba} to embed trajectory features efficiently \cite{suo2024mamba}, modeling contextual information within sequences while maintaining low computational cost.

To process trajectory data effectively, we first organize the data into a unified form that captures its temporal features. We apply average padding to preprocess the input and consolidate trajectory data \(Tr\) into temporal features, denoted as \(X^{\mathbb{T}} \in \mathbb{R}^{N \times M_{\text{max}} \times F}\), where \(N\) is the number of trajectories, \(M_{\text{max}}\) is the maximum sequence length, and \(F\) is the embedding dimension of the temporal features.

Using Mamba, we process these temporal features to capture long-term dependencies in the time series. The Mamba model is based on two key equations: the state equation and the observation equation. 

The state equation models the evolution of the hidden state \(h(t)\) over time:
\begin{equation}
h'(t) = A h(t) + B \mathcal{I}(t),
\end{equation}
where \(A \in \mathbb{R}^{N \times N}\) and \(B \in \mathbb{R}^{N \times 1}\) are the state transition and control matrices, and \(\mathcal{I}(t)\) represents the input at time \(t\).

The observation equation describes how observed outputs are derived from the hidden state:
\begin{equation}\label{e2}
\mathcal{O}(t) = C h(t),
\end{equation}
where \(C \in \mathbb{R}^{N \times 1}\) is the output matrix.

To calculate the output sequence \(y(t)\) at a given time \(t\), we discretize the model using a time scale factor \(\Delta\) for simplification. Mamba applies the zero-order hold (ZOH) method \cite{gu2023mamba} to discretize the matrices, yielding:
\begin{equation}
\begin{aligned}
\bar{A} &= \exp(\Delta A), \\ 
\bar{B} &= \Delta A^{-1} (\exp(\Delta A) - I) \cdot \Delta B, \\ 
\bar{C} &= C.
\end{aligned}
\end{equation}

The discretized form of the state space model is then expressed as:
\begin{equation}
\begin{aligned}
h_{t+1} &= \bar{A} h_t + \bar{B} \, \mathcal{I}_t, \\ 
\mathcal{O}_t &= \bar{C} h_t.
\end{aligned}
\end{equation}

Unlike traditional SSMs, selected SSM in Mamba treats \(\Delta\), \(B\), and \(C\) as learnable parameters, enhancing the model's ability to adapt and learn effectively. As described in \cite{gu2023mamba}, Mamba uses linear layers to process \(\Delta\), \(B\), and \(C\), with \(A\) remaining fixed, which simplifies computation:
\begin{equation}
\begin{aligned}
h_{t+1} &= \exp(\Delta A) h_t + \Delta B \, \mathcal{I}_t, \\ 
\mathcal{O}_t &= C h_t.
\end{aligned}
\end{equation}

The Mamba architecture in our TLM includes the selected SSM, multiple multilayer perceptrons (MLPs), and convolutional blocks, all utilizing the SiLU activation function and nonlinear transformations, as shown in Figure \ref{fig1}. The temporal features \(X^{\mathbb{T}}\) are passed through Mamba to obtain the final temporal embedding \(E^{\mathbb{T}} \in \mathbb{R}^{N \times M_{\text{max}} \times F}\).

\subsection{Attribute Extraction Module with Feature Engineering}

In this study, we introduce the Attribute Extraction Module (AEM), a feature engineering solution designed to transform the sequential features of truck trajectories into rich attribute features. These attribute features offer a generalized description of the overall distribution and key features of the trajectory data, helping the model capture a broader range of information and enhance its generalization ability when facing sparse data or variable-length sequences.

Specifically, by calculating statistical variables—including maximum, minimum, mean, and variance—of differences, ranges, central positions, and angles for the GPS trajectory features \( [t_i, \lambda_i, \phi_i, v_i, \theta_i, e_i] \), the AEM delves into the details of trajectory data. This process reduces data sparsity by summarizing variable-length sequences into fixed-size attribute vectors, and simplifies data complexity by condensing high-dimensional sequential data into lower-dimensional representations. 

\textbf{(a) GPS time:} Regarding the temporal information, we employ the time difference and rate of change to reflect the temporal patterns and trends.

For each trajectory data $G_n$, the time difference between adjacent time points $t_i$ and $t_{i-1}$ can be expressed as:
\begin{equation}
    TimeDiff_{i,n} = t_{i,n} - t_{i-1,n}, \quad i = 2, 3, ..., M_{max}.
\end{equation}

The time rate of change represents the rate of time change between adjacent time points and can be expressed as:
\begin{equation}
TimeRate_{i,n} = \frac{TimeDiff_{i,n}}{t_{i-1,n}}, \quad i = 2, 3, ..., M_{max}.
\end{equation}

\textbf{(b) GPS longitude and latitude:} Concerning the latitude and longitude information, we employ the range and central position of both longitude and latitude to describe the truck's movement range and spatial distribution.

For each trajectory data $Tr_n$, the range of longitude and latitude can be expressed as:
\begin{equation}
\begin{aligned}
LongitudeRange_n &= max(\lambda_{n}) - min(\lambda_{n})\\
LatitudeRange_n &= max(\phi_{n}) - min(\phi_{n}).
\end{aligned}
\end{equation}

For each trajectory data $Tr_n$, the central position of longitude and latitude can be expressed as:
\begin{equation}
\begin{aligned}
LongitudeCenter_n &= \frac{1}{N_n} \sum_{i=1}^{N_n} \lambda_{i,n}\\
LatitudeCenter_n &= \frac{1}{N_n} \sum_{i=1}^{N_n} \phi_{i,n}.
\end{aligned}
\end{equation}

\textbf{(c) Speed:} In terms of speed information, we employ statistical features of speed and speed rate of change to reveal the truck's movement speed and variations.

For each trajectory data $Tr_n$, we calculated the speed at time $t_i$ as follows:
\begin{equation}
Speed_{i,n} = v_{i,n}, i = 1, 2, ..., M_{max}.
\end{equation}
The speed rate of change indicates the rate of speed change between adjacent time points and can be expressed as:
\begin{equation}
SpeedRate_{i,n} = \frac{Speed_{i,n} - Speed_{i-1,n}}{TimeDiff_{i,n}}, \quad i = 3, 4, ..., M_{max}.
\end{equation}

\textbf{(d) Direction:} When dealing with directional information, we employ the rate of change of direction and discretized direction angles to capture statistical features of the direction, enabling a better understanding of the truck's movement direction and variations.

For each trajectory data $Tr_n$, the direction difference can be expressed as:
\begin{equation}
DirectionDiff_{i,n} = \theta_{i,n} - \theta_{i-1,n}, \quad i = 2, 3, ..., M_{max}.
\end{equation}

To discretize the direction angles, we transform continuous direction angle values into discrete categories based on a predefined partition interval $[b_1, b_2, b_3, \ldots, b_{k-1}]$, such as $[0, 45, 90, \ldots, 360]$, where each category spans 45 degrees. By assigning values to corresponding labels, the discretized direction angle can be expressed as:
\begin{equation}
DirectionAngle_{i,n}= \begin{cases}0, & \text { if} \quad \theta_{i,n}<b_1 \\ 1, & \text { if} \quad b_1 \leq \theta_{i,n}<b_2 \\ \vdots & \\ k-1, & \text { if} \quad \theta_{i,n} \geq b_{k-1}\end{cases}.
\end{equation}

\textbf{(e) Event:}  In the context of truck driving, an event typically refers to a specific occurrence or condition that is noteworthy during the vehicle's operation. These events can range from routine operations to anomalies and are categorized as follows:

Normal Operation (0): GPS data is recorded accurately, with the truck operating normally.  
Longitude Error (1): Significant deviation in longitude data, possibly due to signal interference or hardware issues.  
Latitude Error (2): Abnormal latitude data affecting position accuracy, potentially from similar causes as longitude errors.  
Time Error (3): Irregularities in timestamp data, such as jumps, likely due to synchronization or GPS device issues.  
Speed Error (4): Unrealistic speed data (sudden spikes or drops) possibly from sensor glitches.  
Direction Error (5): Inconsistent heading data, misaligned with the truck's path, possibly due to compass errors or signal disruptions.
Given this categorization, the difference in the truck's driving events can be expressed as:

\begin{equation} EventDiff_{i,n} = e_i - e_{i-1}, \quad i = 2, 3, ..., M_{\text{max}}. \end{equation}

\begin{figure}[h!]
\centering
\includegraphics[width=0.47\textwidth]{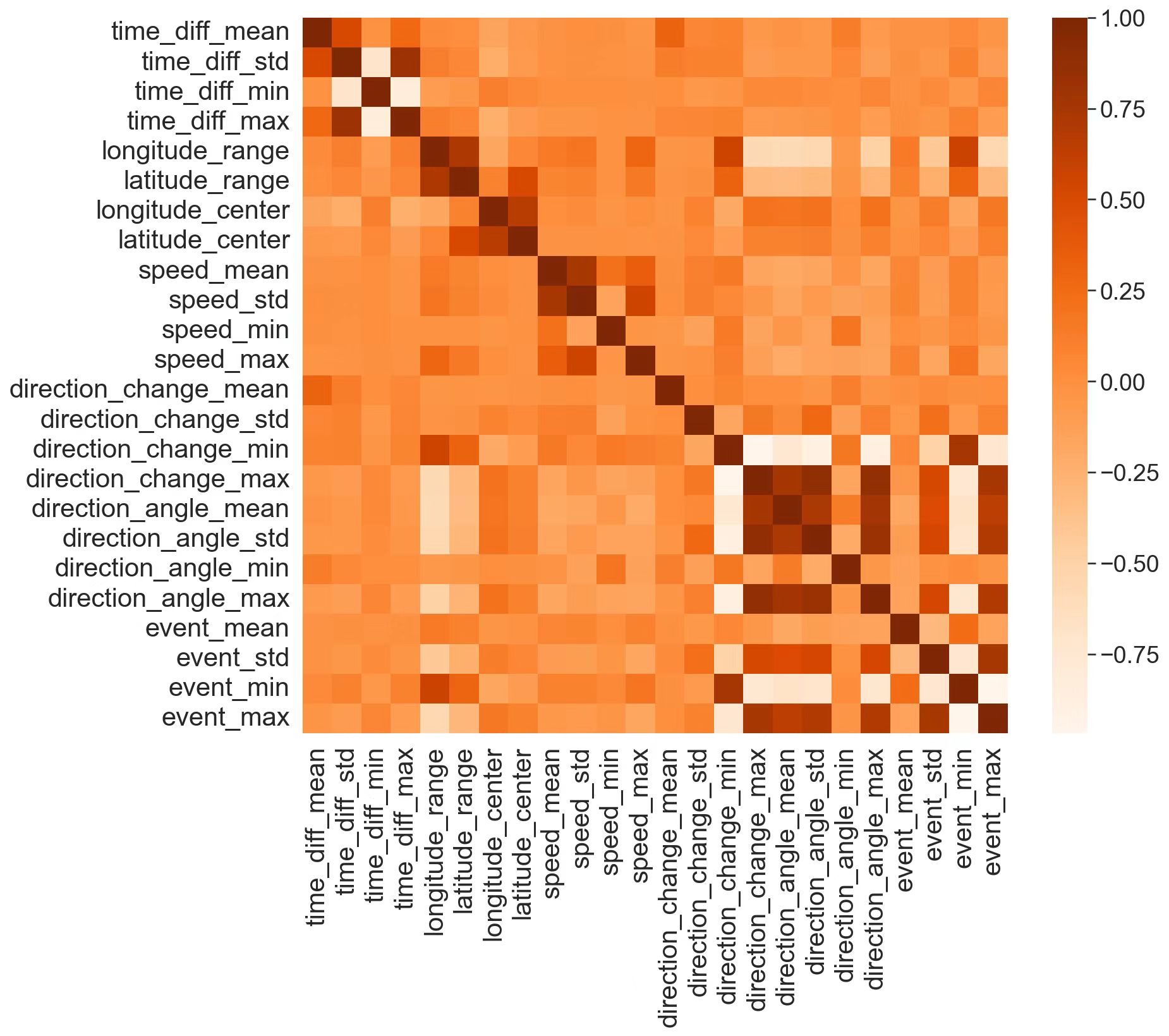}
\caption{Heat-map of attribute embedding correlation.}
\label{fig2}
\end{figure}

Finally, the mean, variance, maximum, and minimum values are computed along the time dimension for all features except longitude and latitude (i.e., \( TimeDiff_{2,n} \), \( TimeRate_{2,n} \), \( Speed_{2,n} \), \( SpeedRate_{3,n} \), \( DirectionRate_{3,n} \), and \( DirectionAngle_{2,n} \)). Then, the attribute embedding \( E^{\mathbb{A}}_n \) for each trajectory data \( Tr_n \) can be obtained:

\begin{equation}
\begin{split}
E^{\mathbb{A}}_n = [&TimeDiff_{n},\ TimeRate_{n}, \\
&LongitudeRange_n,\ LatitudeRange_n, \\
&LongitudeCenter_n, LatitudeCenter_n,\\
& Speed_{n},\ SpeedRate_{n}, \\
&DirectionDiff_{n},\ DirectionAngle_{n}, \\
&EventDiff_n].
\end{split}
\end{equation}

We obtain the attribute embedding \( E^{\mathbb{A}} \in \mathbb{R}^{N \times D^a} \) for all trajectory data, where \( D^a = 24 \) is the dimensionality of the attribute feature space and satisfies \( D^a \ll M_{\text{max}} \times F \). By reducing the dimensionality from the potentially large \( M_{\text{max}} \times F \) to a fixed \( D^a \), we effectively reduce data sparsity and complexity, making the data more manageable for the model. Figure \ref{fig2} presents the correlation graph among attribute features, providing insights into their interactions and relationships.

\subsection{Spatial Fusion Module with Feature Diffusion Graph Learning}
The purpose of the Spatial Fusion Module (SFM) is to mine the intrinsic influence relationships caused by the collaborative operation of different trajectories based on a spatiotemporal relation graph, and to coordinate the three feature spaces of time, attributes, and spatial dimensions. Firstly, SFM relies on the temporal sequence embedding $E^\mathbb{T}$ to construct a spatiotemporal relation graph $G$ based on sequence features, aiming to capture the complex spatiotemporal correlations between trajectories. Secondly, it carefully considers the synchronous impact of the spatiotemporal correlation graph and attribute embedding $E^\mathbb{A}$ using a feature diffusion graph learning method.

\subsubsection{Spatio-temporal Relations Graph}

In the AEM, the detailed transformation of attribute data may inadvertently lead to varying degrees of information loss. To intelligently preserve as much temporal information as possible, we construct a spatiotemporal relation graph $G$ based on the similarity of temporal features. By employing the nearest neighbor algorithm, we establish an adjacency matrix based on the similarity of temporal features, represented as the spatial embedding $E^\mathbb{S} \in \mathbb{R}^{N \times N}$. Specifically, we first flatten the temporal sequence embedding of trajectories $E^\mathbb{T} \in \mathbb{R}^{N \times M_{max} \times F}$ to obtain a matrix of shape $\mathbb{R}^{N \times (M_{max} * F)}$, then calculate the similarity measure (such as using Euclidean distance or Manhattan distance) between all pairs of trajectory samples. Based on the results of these similarity calculations, we select $K$ most similar trajectory nodes as neighbors for each target node, thus forming the spatiotemporal relation graph. Here, we use Euclidean distance as an example to illustrate the measurement of similarity.

First, we construct the Euclidean distance matrix $\mathbb E$ by calculating the Euclidean distance between two temporal feature vectors $E^\mathbb{T}_i$ and $E^\mathbb{T}_j$:

\begin{equation}
\mathbb E_{i,j} = \sqrt{\sum_{k=1}^{M_{max} * F} (E^\mathbb{T}_{i,k} - E^\mathbb{T}_{j,k})^2},
\end{equation}
where $E^\mathbb{T}_{i,k}$ and $E^\mathbb{T}_{j,k}$ represent the temporal sequence embedding $E^\mathbb{T}_i$ and $E^\mathbb{T}_j$ at the $k^{th}$ feature dimension, respectively, and $M_{max} * F$ is the total dimensionality of the feature vector.

Next, for each trajectory sample $i$, we calculate its $K$ nearest neighbors based on the Euclidean distance:

\begin{equation}
\mathrm{Neighbors}_K(i) = \underset{j \neq i}{\mathrm{argmin}K},  \mathbb E_{i,j},
\end{equation}
where $\mathrm{Neighbors}_K(i)$ represents the index set of $i$'s $K$ nearest neighbors, and $\underset{j \neq i}{\mathrm{argmin}K}$ denotes the indices of $K$ samples $j$ for which $\mathbb E_{i,j}$ is minimized, excluding the sample itself.

We then construct a similarity adjacency matrix $W$, where nodes represent trajectory samples and edges represent the spatiotemporal relations between trajectories. In $W$, the value of $W_{i,j}$ is $\mathbb E_{i,j}$ if sample $j$ is one of the $K$ nearest neighbors of sample $i$, and 0 otherwise.

Finally, combined with attribute embedding $E^\mathbb{A}$, we construct a spatio-temporal relations graph $G = \{W,E^{\mathbb{A}}\}$.

\subsubsection{Feature Diffusion Graph Learning}

In addition, we employ a feature diffusion (FD) method to achieve the diffusion and aggregation of attribute embedding. Given the graph data, we appily adjacency matrix $W$ to update the attribute embedding $E^{\mathbb{A}}$, facilitating the exchange of information and aggregation of features among nodes. The mathematical description of the message passing is as follows:
\begin{equation}
 {E^\mathbb{S}}^{(l+1)} = \begin{cases} W {E^\mathbb{A}}^{(l)} & \text { if} l=0\\
W {E^\mathbb{S}}^{(l)} & \text { if} l\neq 0
 \end{cases},
\end{equation}
where ${E^{\mathbb{S}}}^{(l+1)}$ represents the spatial embedding after the $l$-th iteration. In the first iteration, the adjacency matrix $W$ is multiplied with the current spatial embedding $E^{\mathbb{A}}$ to obtain the updated spatial embedding ${E^{\mathbb{S}}}^{(l +1)}$. Through multiple iterations of propagation, the spatial embeddings gradually aggregate and fuse, capturing global and local information in the graph structure.

Normalization is applied to the adjacency matrix $W$, resulting in the normalized adjacency matrix ${D}^{-1/2}{WD}^{-1/2}$, where $D$ is the degree matrix \cite{kipf2016semi}. Normalization helps balance the weights of different nodes in the graph, thus promoting more equitable and stable feature propagation. Ultimately, the propagation process can be expressed as:
\begin{equation}
 {E^\mathbb{S}}^{(l+1)} = \begin{cases} {D}^{-1/2}{WD}^{-1/2} {E^\mathbb{A}}^{(l)} & \text { if} l=0\\
{D}^{-1/2}{WD}^{-1/2} {E^\mathbb{S}}^{(l)} & \text { if} l\neq 0
 \end{cases}.
\end{equation}
By specifying the iteration parameter $l$, we can adjust the extent of feature propagation. Smaller values of $l$ are better suited for capturing local information, while larger ones are advantageous for incorporating global information.

\subsection{Optimize and Train for ETA}
Through the three modules of TLM, AEM and SFM, we obtain an embedding that represents the three dimensions of time series, attribute and space. Next, we introduce framework loss function training and how to perform arrival time estimation by leveraging DPM.
\subsubsection{Optimize Loss Training}
To optimize framework learning, we designed two loss functions for embedding learning and structural learning optimization. 

The loss function for embedding learning is represented as:
\begin{equation}
\mathcal{L}_\text{E} = \frac{1}{N}\sum_{i=1}^{N} \left( \frac{E^\mathbb{T}_i \cdot X^\mathbb{T}_i}{|E^\mathbb{T}_i| |X^\mathbb{T}_i|} - \frac{E^\mathbb{T}_j \cdot X^\mathbb{T}_j}{|E^\mathbb{T}_j| |X^\mathbb{T}_j|} \right)^2.
\end{equation}
The embedding learning loss focuses on optimizing the temporal sequence embeddings $E^\mathbb{T}$ to ensure that the similarity between embeddings reflects the similarity between the original temporal sequences. By comparing the similarity (using cosine similarity as the measure) between the embedding $E^\mathbb{T}$ and its corresponding original temporal sequence embedding $X^\mathbb{T}$, $\mathcal{L}_\text{E}$ encourages the model to learn embeddings that retain the intrinsic features of the original temporal sequences.

For structural learning, the loss function can be represented as:
\begin{equation}
\mathcal{L}_\text{S} = \frac{1}{N}\sum_{i=1}^{N} \left((E^\mathbb{T}_i)^{\top}\sum_{j=1}^{N}\left(\sqrt{\sum_{k=1}^{D^a} (E^\mathbb{S}_{i,k} - E^\mathbb{S}_{j,k})^2}\right)E^\mathbb{T}_i \right).
\end{equation}
The structural learning loss function further considers the relationship between  features $E^\mathbb{S}$ and temporal sequence embeddings $E^\mathbb{T}$. By integrating the temporal embeddings with a weighted distance based on the difference in attribute embeddings, $\mathcal{L}_\text{S}$ aims to optimize the structure of the temporal sequence embeddings to better reflect the intrinsic structure and relationships within the trajectory data.

Combining both losses, the total loss is expressed as:
\begin{equation}
 \mathcal{L}_\text{SE} =\mathcal{L}_\text{E} + \eta \mathcal{L}_\text{S},
\end{equation}
where $\eta$ is a tuning parameter. This self-supervised approach to learning temporal embeddings leverages inherent data properties and relationships without requiring external labels for supervision. By optimizing $\mathcal{L}_\text{SE}$, we facilitate the model to autonomously discover and encode meaningful patterns and structures from the time series data, enhancing its ability to generate embeddings that are reflective of the underlying temporal dynamics and interactions. 

\subsubsection{Downstream Prediction Module}
Considering the earlier discussion, let's define the total loss function for learning temporal embeddings in a self-supervised manner. To estimate the arrival time, we first fuse the embeddings representing time, attributes, and space using residual connections \cite{he2016deep} to obtain a hybrid embedding:
\begin{equation}
E^\mathbb{H} = {E^\mathbb{A}} + \alpha {E^\mathbb{S}},
\end{equation}
where $\alpha$ represents the balance parameter. {The addition is crucial as it enables the model to utilize rich attribute information alongside dynamic spatial features, enhancing the overall representation.}

Then, we input the hybrid embedding $E^\mathbb{H}$ into Histogram-based Gradient Boosting (HGB) \cite{ke2017lightgbm} as a predictor for estimating arrival times. The core idea of HGB is to utilize the gradient boosting framework combined with histogram technology to optimize the training process of decision trees. HGB constructs a series of decision trees to gradually approximate the target function, with each tree learning the direction of the residual predicted by the previous tree, thereby enhancing the model's predictive capability. Through HGB, we map the hybrid embedding $E^\mathbb{H}$ to the true arrival time $\hat{Y}$. If it is a regression task, the loss function can be represented as:

\begin{equation}
\mathcal{L}_{\text{HGB}} = \frac{1}{N}\sum_{i=1}^{N} (Y_i - \hat{Y}_i)^2,
\end{equation}
where $Y_i$ is the actual arrival time of the $i^{th}$ sample, $\hat{Y}_i$ is the corresponding predicted arrival time, and $N$ is the total number of samples.

Overall, our framework cleverly avoids directly inputting the high-dimensional temporal embedding $E^\mathbb{T}$ into the predictor, significantly reducing computational complexity. By combining the attribute embedding with the last layer's embedding, we effectively merge information from multiple sources to produce a comprehensive hybrid representation. The attribute embedding contains the initial features of the input data, carrying key information, while the final layer's embedding, obtained through multiple rounds of propagation and aggregation, provides advanced representations in time and space.

\subsection{Computational Complexity Analysis}

In this section, we analyze the temporal and spatial complexity of three modules.

\textbf{TLM}:
Assuming the truck trajectory dataset comprises $N$ samples, with each sample including data from $M$ time instances, the literature \cite{gu2023mamba} indicates that the TLM, when employing the Mamba model, can achieve a computational complexity of merely $O(N)$.

\textbf{AEM}:
The AEM primarily involves extensive statistical analysis and feature extraction from truck trajectory data. This includes computations of time differences, time rate of change, longitude and latitude ranges, centers, speeds, speed rate of change, and direction rate of change. The complexity for calculating time differences and time rate of change is $O(M)$ per sample, computing longitude and latitude range and centers is $O(M)$ per sample, calculating speeds and speed rate of change is $O(M)$ per sample, and determining direction rate of change is $O(M)$ per sample. Consequently, the total complexity of the AEM module is $O(NM)$.

\textbf{SFM}:
The SFM involves computing similarity adjacency matrices and feature propagation in graph learning. For calculating similarity adjacency matrices, we utilize the Nearest Neighbors algorithm with a complexity of $O(NM^2)$, where $N$ is the number of samples and $M$ is the number of time instances. Feature propagation in graph learning mainly revolves around iterative updates of the adjacency matrix. Assuming $l$ iterations, each iteration's computational complexity is $O(NM^2)$, leading to a total complexity of $O(lNM^2)$ for feature propagation in graph learning. Thus, the overall complexity of the SFM is $O(lNM^2)$.

\section{Experiments and Analysis}\label{s4}
In this section, we conduct a comprehensive experimental evaluation of our proposed framework to validate its effectiveness and performance in estimating truck arrival times.

\begin{table}[htbp]\scriptsize
  \centering
  \caption{Data statistics.}
    \begin{tabular}{cccccc}
    \toprule
          & $N$     & $M_{min}$ & $M_{max}$ & $M_{mean}$ & $M_{std}$ \\
    \midrule
    Baoan & 1538  & 2     & 1997  & 613.66  & 612.19  \\
    Nanshan & 1062  & 2     & 1998  & 580.39  & 618.20  \\
    Yantian & 2899  & 1     & 1999  & 918.50  & 592.89  \\
    Futian & 716   & 2     & 1995  & 703.30  & 672.64  \\
    Luohu & 272   & 2     & 1978  & 546.84  & 558.35  \\
    \midrule
    All   & 6487  & 1     & 1999  & 672.54  & 610.85  \\
    \bottomrule
    \end{tabular}%
  \label{tab1}%
\end{table}%

\begin{figure*}[t]
\centering
\includegraphics[width=0.7\textwidth]{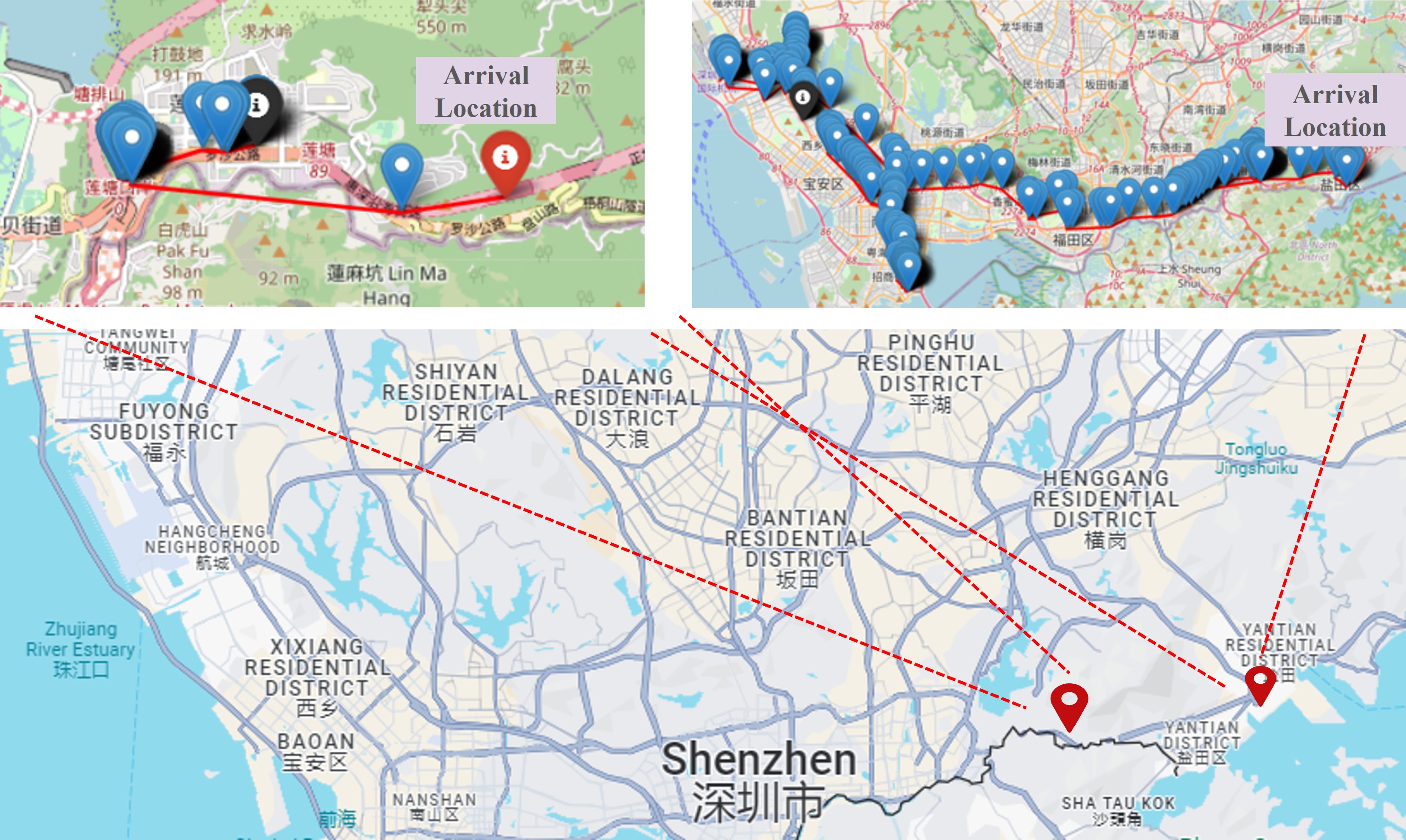}
\caption{{Illustration of GPS track recording locations of trucks in Shenzhen.}}
\label{fig4}
\end{figure*}

\subsection{Datasets and Settings}
\subsubsection{Datasets}

\begin{figure*}[t]
\centering
\subfloat[Baoan.]{
\includegraphics[width=0.25\textwidth]{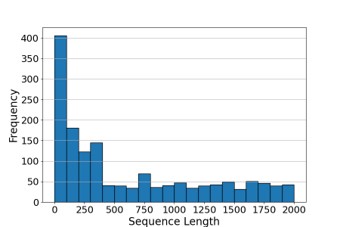}
}
\centering
\subfloat[Nanshan.]{
\includegraphics[width=0.25\textwidth]{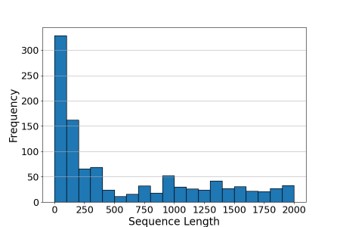}
}
\centering
\subfloat[Yantian.]{
\includegraphics[width=0.25\textwidth]{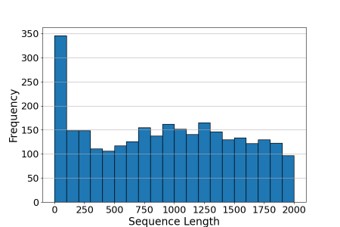}
}

\centering
\subfloat[Futian.]{
\includegraphics[width=0.25\textwidth]{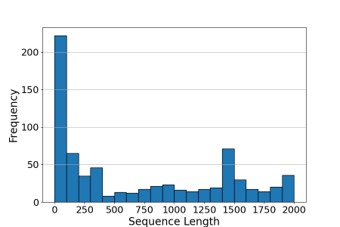}
}
\centering
\subfloat[Luohu.]{
\includegraphics[width=0.25\textwidth]{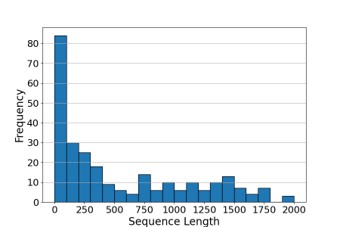}
}
\centering
\subfloat[All.]{
\includegraphics[width=0.25\textwidth]{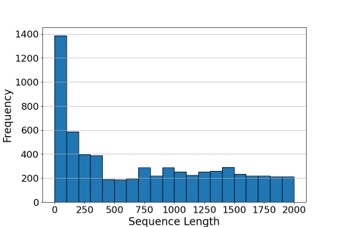}
}
\caption{GPS track and statistics of trucks in Shenzhen.}
\label{fig5}
\end{figure*}

Our dataset was collected from GPS devices installed on a fleet of trucks operating across various districts in Shenzhen, China. Each truck was equipped with a GPS sensor that recorded multiple parameters at regular intervals, including time, geographic coordinates (longitude and latitude), speed, direction, and specific event information. To validate our framework, we gathered truck trajectory data from five districts in Shenzhen: Baoan, Nanshan, Yantian, Futian, and Luohu. This dataset contains a total of 3,622,017 data points, forming 6,487 trajectories (as illustrated in Figure \ref{fig4}). Table \ref{tab1} provides a detailed breakdown of trajectory data across these five regions, including the number of trajectories ($N$), minimum length ($M_{min}$), maximum length ($M_{max}$), average length ($M_{mean}$), and standard deviation ($M_{std}$) of each trajectory. The results show that trajectory lengths are generally distributed between 500 and 900 data points, with a standard deviation between 500 and 600, highlighting the sparsity and variability of trajectory lengths.
To more intuitively illustrate the features of our dataset, we visualized the distribution of trajectory lengths across different regions (as shown in Figure \ref{fig5}). The figure demonstrates a long-tail distribution of trajectory lengths, with most trajectories containing fewer than 250 data points, while a few extend between 250 and 2,000 data points. This sparsity and imbalance in sequence length present significant challenges for accurately predicting truck arrival times.

\subsubsection{Baselines}

To evaluate the performance of TAS-TsC, we select several popular methods as baselines:
\begin{itemize}
\item Attribute-based Models: Linear Regression (LR) \cite{yu2011bus}, Histogram Gradient Boosting (HGB) \cite{zhang2015gradient} and Extreme Gradient Boosting (XGB) \cite{shi2021prediction}, and STNN\cite{jindal2017unified}.
\item Graph-Based Models: Graph Convolutional Network (GCN) \cite{kipf2016semi}, and Graph Attention Network (GAT) \cite{velivckovic2017graph}.
\item Sequence Models: Historical Averages (HA) \cite{chen2011historical},  Recurrent Neural Networks (RNN) \cite{connor1994recurrent}, Long Short-Term Memory (LSTM) \cite{hochreiter1997long}, and Gated Recurrent Units (GRU) \cite{noman2021towards}, and BiLSTM \cite{hochreiter1997long}.
\item Hybrid-Based Models: MetaTTE \cite{wang2022fine} and Inductive Graph Transformers (IGT) \cite{zhou2023inductive}.
\end{itemize}
In comparing our framework against the baseline methods, we adhere to the default parameters for all methods while acknowledging that some of the methods are not entirely compatible with our data input format in their original design. We perform feature engineering on our data and make minor modifications to the baseline models, ensuring these adjustments do not impact the overall integrity of the comparison. For instance, for IGT, we configure the features according to the original GPS data of our dataset and construct heterogeneous graphs based on different district boundaries.

\subsubsection{Experimental Setup}
All experiments are implemented on PyTorch platform with an NVIDIA 4090 GPU. The maximum Epoch is 100, using the Adam \cite{kingma2014adam} optimizer with a learning rate of 1e-4.
For each method, strict parameter tuning is performed to ensure optimal task performance. The data is performed at a ratio of 7:1:2 for training, validation and testing, respectively.

To gauge the efficacy of our models, we employ a suite of metrics \cite{shcherbakov2013survey} renowned for their ability to capture the essence of prediction accuracy and model robustness:
\begin{itemize}
 \item Mean Squared Error (MSE): A measure of the average squared differences between predicted and actual values.
\item {Root Mean Squared Error (RMSE)}: It captures the square root of the MSE, providing an interpretation of error in the original units.
\item {Mean Absolute Percentage Error (MAPE)}: This metric presents prediction error as a percentage, enabling a relative understanding of the model's performance.
\item {Mean Absolute Error (MAE): Measures the average absolute differences between predicted and actual values, representing error in the same units as the data.}
\end{itemize}

\subsection{Comparison of Estimating Time of Arrival}

Estimating truck arrival times accurately based on historical trajectory data poses a challenging regression problem, especially due to the complex, sparse, and variable-length nature of the data. In this section, we evaluate our proposed framework alongside various comparative models across four main categories: attribute-based, graph-based, sequence-based, and hybrid prediction methods.

Table \ref{tab_a} presents the results across three evaluation metrics (MSE, RMSE, and MAPE) on five individual datasets and a composite dataset (``All") combining all five. Our results reveal that while attribute-based and sequence-based models demonstrate reasonable accuracy on MSE and RMSE, they lack robustness in capturing the intricate spatial dependencies inherent in truck trajectories. Graph-based models perform relatively well on MAPE, indicating their strength in handling irregular data points. However, the hybrid model, which integrates multiple embedding representations, consistently outperforms all other categories across all metrics. This performance boost highlights the advantages of combining temporal, spatial, and attribute-based information to capture both localized and broad contextual factors influencing arrival times.

Our TAS-TsC framework, which enables dynamic interactions among temporal, attribute, and spatial modules, achieves the best performance, demonstrating an 8.7\%, 14.1\%, 11.2\%, and 23.6\% improvement over the second-best baseline (IGT) on MSE, RMSE, and MAPE in the ``All" dataset, respectively. By leveraging hybrid embeddings, our method effectively addresses the challenges of sparse and variable-length trajectories, enabling more precise modeling of truck arrival times. This comprehensive integration proves crucial for extracting relevant temporal patterns and spatial relationships, which are otherwise overlooked by single-method approaches.

\begin{table*}[h]\scriptsize
  \centering
  \caption{{The comparative results of various time-series methods on arrival time estimation. Bold black font indicates the best results.}}
    \renewcommand\arraystretch{0.8}
   \renewcommand\tabcolsep{4pt}
    \begin{tabular}{cccccccccccccc}
    \toprule
\multirow{2}[4]{*}{\textbf{Type}} & \multirow{2}[4]{*}{\textbf{Methods}} &\multicolumn{4}{c}{\textbf{Baoan}} & \multicolumn{4}{c}{\textbf{Nanshan}} & \multicolumn{4}{c}{\textbf{Yantian}} \\
\cmidrule{3-14}          &       & \textbf{MSE} & \textbf{RMSE} & \textbf{MAPE} & \textbf{MAE} & \textbf{MSE} & \textbf{RMSE} & \textbf{MAPE} & \textbf{MAE} & \textbf{MSE} & \textbf{RMSE} & \textbf{MAPE} & \textbf{MAE} \\
    \midrule
    \multirow{4}[1]{*}{Attribute} & LR    & 0.0802  & 0.2832  & 4.0516  & 0.2455  & 0.0832  & 0.2885  & 8.4107  & 0.2612  & 0.8982  & 0.9477  & 1.4556  & 0.2396  \\
          & HGB   & 0.0296  & 0.1721  & 1.6093  & 0.1883  & 0.0270  & 0.1642  & 0.9412  & 0.2480  & 0.2031  & 0.4506  & 1.5997  & 0.1407  \\
          & XGB   & 0.0312  & 0.1768  & 2.4474  & 0.1857  & 0.0264  & 0.1623  & 0.8410  & 0.2064  & 0.2406  & 0.4905  & 1.8829  & 0.2181  \\
          & STNN  & 0.0763  & 0.2762  & 2.1947  & 0.2350  & 0.1155  & 0.3399  & 1.4642  & 0.2477  & 0.4399  & 0.6633  & 2.1455  & 0.2214  \\
    \midrule
    \multirow{2}[2]{*}{Graph} & GCN   & 0.1402  & 0.3889  & 0.6788  & 0.1499  & 0.1505  & 0.4382  & 0.6760  & 0.1938  & 0.0994  & 0.3397  & 0.9459  & 0.1310  \\
          & GAT   & 0.1146  & 0.3253  & 0.5194  & 0.1801  & 0.1389  & 0.3491  & 0.5979  & 0.1846  & 0.1144  & 0.3255  & 0.8915  & 0.1346  \\
    \midrule
    \multirow{5}[2]{*}{Sequence} & HA    & 0.0837  & 0.2894  & 3.9405  & 0.2316  & 0.0885  & 0.2976  & 7.7419  & 0.1915  & 1.0565  & 1.0279  & 1.0260  & 0.2181  \\
          & RNN   & 0.0663  & 0.2574  & 1.1395  & 0.1960  & 0.0632  & 0.2515  & 1.2122  & 0.1643  & 0.7715  & 0.8783  & 2.5798  & 0.1347  \\
          & LSTM  & 0.0647  & 0.2544  & 0.8745  & 0.1829  & 0.0620  & 0.2489  & 1.1458  & 0.1657  & 0.6429  & 0.8018  & 2.2278  & 0.1435  \\
          & GRU   & 0.0651  & 0.2551  & 0.9536  & 0.1841  & 0.0854  & 0.2923  & 1.1199  & 0.1237  & 0.4973  & 0.7052  & 1.7717  & 0.1285  \\
          & BiLSTM & 0.0685  & 0.2617  & 0.5307  & 0.1962  & 0.0805  & 0.2837  & 0.7548  & 0.1492  & 0.3191  & 0.2878  & 1.0301  & 0.1001  \\
    \midrule
    \multirow{3}[2]{*}{Hybrid} & MetaTTE & 0.0316  & 0.1778  & 0.4768  & 0.1194  & 0.0623  & 0.2496  & 0.6018  & 0.1100  & 0.0779  & 0.2791  & 0.9346  & 0.0969  \\
          & IGT   & 0.0312  & 0.1768  & 0.4958  & 0.1168  & 0.0182  & 0.1351  & 0.6865  & 0.1004  & 0.0466  & 0.2158  & 0.8404  & 0.0922  \\
          & Ours  & \textbf{0.0199} & \textbf{0.1410} & \textbf{0.3509} & \textbf{0.0964} & \textbf{0.0147} & \textbf{0.1212} & \textbf{0.4696} & \textbf{0.0944} & \textbf{0.0136} & \textbf{0.1168} & \textbf{0.6826} & \textbf{0.0868} \\
    \midrule
\multirow{2}[4]{*}{\textbf{Type}} & \multirow{2}[4]{*}{\textbf{Methods}} & \multicolumn{4}{c}{\textbf{Futian}} & \multicolumn{4}{c}{\textbf{Luohu}} & \multicolumn{4}{c}{\textbf{All}} \\
\cmidrule{3-14}          &       & \textbf{MSE} & \textbf{RMSE} & \textbf{MAPE} & \textbf{MAE} & \textbf{MSE} & \textbf{RMSE} & \textbf{MAPE} & \textbf{MAE} & \textbf{MSE} & \textbf{RMSE} & \textbf{MAPE} & \textbf{MAE} \\
    \midrule
    \multirow{4}[2]{*}{Attribute} & LR    & 0.0861  & 0.2934  & 1.1512  & 0.2638  & 0.1543  & 0.3928  & 1.2551  & 0.2638  & 0.0864  & 0.2939  & 2.4803  & 0.2645  \\
          & HGB   & 0.0251  & 0.1585  & 0.3945  & 0.2534  & 0.0416  & 0.2040  & 0.4220  & 0.1409  & 0.0416  & 0.2040  & 0.7598  & 0.2563  \\
          & XGB   & 0.0297  & 0.1722  & 0.3485  & 0.2509  & 0.0729  & 0.2701  & 0.8608  & 0.1774  & 0.0241  & 0.1551  & 0.6950  & 0.1741  \\
          & STNN  & 0.0517  & 0.2275  & 0.7228  & 0.2598  & 0.1293  & 0.3596  & 1.2337  & 0.2509  & 0.0500  & 0.2236  & 1.7423  & 0.2554  \\
    \midrule
    \multirow{2}[2]{*}{Graph} & GCN   & 0.1580  & 0.3937  & 1.0349  & 0.1774  & 0.1559  & 0.3688  & 0.4779  & 0.1774  & 0.1181  & 0.3360  & 0.4395  & 0.1752  \\
          & GAT   & 0.1520  & 0.4078  & 1.0421  & 0.1774  & 0.1258  & 0.3363  & 0.5165  & 0.1826  & 0.1236  & 0.3263  & 0.3980  & 0.1696  \\
    \midrule
    \multirow{5}[1]{*}{Sequence} & HA    & 0.0962  & 0.3102  & 1.3157  & 0.1382  & 0.0933  & 0.3055  & 1.1704  & 0.1774  & 0.0933  & 0.3055  & 2.7137  & 0.2455  \\
          & RNN   & 0.0673  & 0.2595  & 0.8990  & 0.1508  & 0.4342  & 0.6589  & 0.6027  & 0.1508  & 0.0421  & 0.2052  & 0.9105  & 0.1717  \\
          & LSTM  & 0.0250  & 0.1326  & 0.3069  & 0.1406  & 0.0592  & 0.2433  & 0.5732  & 0.1382  & 0.0517  & 0.2273  & 0.8091  & 0.1625  \\
          & GRU   & 0.0705  & 0.2654  & 0.3177  & 0.1409  & 0.0611  & 0.2472  & 0.4395  & 0.1213  & 0.0453  & 0.2129  & 0.7094  & 0.1635  \\
          & BiLSTM & 0.0713  & 0.2670  & 0.3185  & 0.1508  & 0.0995  & 0.3155  & 1.0395  & 0.1407  & 0.0476  & 0.2182  & 0.7761  & 0.1659  \\
           \midrule
    \multirow{3}[1]{*}{Hybrid} & MetaTTE & 0.0233  & 0.1526  & 0.2823  & 0.1101  & 0.0490  & 0.2213  & 0.3889  & 0.1164  & 0.0275  & 0.1658  & 0.4770  & 0.1587  \\
          & IGT   & 0.0206  & 0.1435  & 0.2910  & 0.1111  & 0.0421  & 0.2052  & 0.3824  & 0.1146  & 0.0254  & 0.1565  & 0.4966  & 0.1328  \\
          & Ours  & \textbf{0.0153} & \textbf{0.1236} & \textbf{0.2387} & \textbf{0.1030} & \textbf{0.0294} & \textbf{0.1713} & \textbf{0.3573} & \textbf{0.1033} & \textbf{0.0181} & \textbf{0.1345} & \textbf{0.4410} & \textbf{0.1014} \\
    \bottomrule
    \end{tabular}%
  \label{tab_a}%
\end{table*}%

\begin{table}[htbp]\scriptsize
  \centering
  \caption{Cross-domain generalization performance results of comparison methods.}
      \renewcommand\arraystretch{0.8}
   \renewcommand\tabcolsep{4pt}
    \begin{tabular}{ccccccc}
    \toprule
    \multirow{2}[4]{*}{\textbf{Methods}} & \multicolumn{3}{c}{\textbf{Baoan-Futian}} & \multicolumn{3}{c}{\textbf{Baoan-Luohu}} \\
\cmidrule{2-7}          & \textbf{MSE} & \textbf{RMSE} & \textbf{MAPE} & \textbf{MSE} & \textbf{RMSE} & \textbf{MAPE} \\
    \midrule
    STNN  & 0.1223  & 0.3497  & 1.0126  & 0.0988  & 0.3143  & 2.0753  \\
    MetaTTE & 0.0965  & 0.3107  & 1.1419  & 0.1507  & 0.3882  & 2.5941  \\
    IGT   & 0.1032  & 0.3212  & 1.1135  & 0.1343  & 0.3665  & 2.3879  \\
    Ours & \textbf{0.0819} & \textbf{0.2862} & \textbf{0.8972} & \textbf{0.0882} & \textbf{0.2970} & \textbf{1.7480} \\
    \midrule
    \multirow{2}[4]{*}{\textbf{Methods}} & \multicolumn{3}{c}{\textbf{Nanshan-Futian}} & \multicolumn{3}{c}{\textbf{Nanshan-Luohu}} \\
\cmidrule{2-7}          & \textbf{MSE} & \textbf{RMSE} & \textbf{MAPE} & \textbf{MSE} & \textbf{RMSE} & \textbf{MAPE} \\
    \midrule
    STNN  & 0.1062  & 0.3259  & 1.5059  & 0.1134  & 0.3368  & 1.4796  \\
    MetaTTE & 0.1064  & 0.3262  & 1.0558  & 0.0999  & 0.3161  & 1.5370  \\
    IGT   & 0.0931  & 0.3052  & 1.3873  & 0.0909  & 0.3015  & 1.4142  \\
    Ours & \textbf{0.0886} & \textbf{0.2977} & \textbf{0.9371} & \textbf{0.0766} & \textbf{0.2768} & \textbf{1.2083} \\
    \bottomrule
    \end{tabular}%
  \label{tab_b}%
\end{table}%

\subsection{Domain Generalization Verification}

Urban logistics environments are characterized by region-specific variations in traffic flow, road structure, and environmental factors, posing significant challenges to model generalizability. In our domain generalization evaluation, we design an experiment that simulates real-world conditions to rigorously test the adaptability of the TAS-TsC framework.

We use data from ``Baoan" and ``Nanshan" as the training regions, while ``Futian" and ``Luohu" are chosen as test regions with distinct traffic and environmental features. This setup allows us to assess the model's capacity to adapt and maintain predictive accuracy in regions not represented in the training data. Our TAS-TsC framework, through its tri-space coordination mechanism, is specifically designed to capture transferable patterns by mining deep correlations between trajectories in the temporal, spatial, and attribute domains.

Table \ref{tab_b} shows that our method achieves superior results across all metrics in this domain transfer setting. This performance is attributed to the model’s ability to leverage comprehensive embeddings that generalize effectively, capturing not only region-specific patterns but also universal features across diverse geographical environments. Our framework’s capacity to adapt to novel traffic conditions and geographical layouts underscores its robustness, making it a valuable tool for real-world applications where deployment may span across multiple, varying urban landscapes.

\begin{table*}[htbp]\scriptsize
  \centering
  \caption{Comparative experimental results of our method under different tmporal sequence prediction models.}
       \renewcommand\arraystretch{0.8}
   \renewcommand\tabcolsep{4pt}
    \begin{tabular}{cccccccccrrrr}
    \toprule
    \multirow{2}[4]{*}{\textbf{Methods}} & \multicolumn{4}{c}{\textbf{Baoan}} & \multicolumn{4}{c}{\textbf{Nanshan}} & \multicolumn{4}{c}{\textbf{Yantian}} \\
\cmidrule{2-13}          & \textbf{MSE} & \textbf{RMSE} & \textbf{MAPE} & \textbf{Time} & \textbf{MSE} & \textbf{RMSE} & \textbf{MAPE} & \textbf{Time} & \multicolumn{1}{c}{\textbf{MSE}} & \multicolumn{1}{c}{\textbf{RMSE}} & \multicolumn{1}{c}{\textbf{MAPE}} & \multicolumn{1}{c}{\textbf{Time}} \\
    \midrule
    RNN   & 0.0428  & 0.2070  & 0.8251  & 101.4300  & 0.0431  & 0.2075  & 0.7143  & 59.1661  & 0.0310  & 0.1760  & 1.7815  & 99.0645  \\
    LSTM  & 0.0479  & 0.2190  & 0.8031  & 72.6494  & 0.0318  & 0.1784  & 0.5308  & 50.0750  & 0.0308  & 0.1755  & 1.5600  & 91.7601  \\
    Transformer & 0.0287  & 0.1695  & 0.5997  & 126.8757  & 0.0273  & 0.1652  & 0.5784  & 56.4320  & 0.0303  & 0.1741  & 1.5915  & 163.7775  \\
    Mamba & \textbf{0.0235} & \textbf{0.1534} & \textbf{0.4188} & \textbf{30.1919} & \textbf{0.0163} & \textbf{0.1278} & \textbf{0.5565} & \textbf{20.6309} & \textbf{0.0280} & \textbf{0.1673} & \textbf{1.4723} & \textbf{69.3556} \\
    \midrule
    \multirow{2}[4]{*}{\textbf{Methods}} & \multicolumn{4}{c}{\textbf{Futian}} & \multicolumn{4}{c}{\textbf{Luohu}} & \multicolumn{4}{c}{\textbf{All}} \\
\cmidrule{2-13}          & \textbf{MSE} & \textbf{RMSE} & \textbf{MAPE} & \textbf{Time} & \textbf{MSE} & \textbf{RMSE} & \textbf{MAPE} & \textbf{Time} & \multicolumn{1}{c}{\textbf{MSE}} & \multicolumn{1}{c}{\textbf{RMSE}} & \multicolumn{1}{c}{\textbf{MAPE}} & \multicolumn{1}{c}{\textbf{Time}} \\
    \midrule
    RNN   & 0.0358  & 0.1892  & 0.4042  & 46.0097  & 0.0641  & 0.2531  & 0.4276  & 17.3720  & 0.0347  & 0.1862  & 0.7050  & 429.8421  \\
    LSTM  & 0.0225  & 0.1500  & 0.2905  & 35.6378  & 0.0534  & 0.2310  & 0.3723  & 12.4982  & 0.0236  & 0.1536  & 0.4386  & 275.9295  \\
    Transformer & 0.0249  & 0.1577  & 0.4166  & 41.3808  & 0.0416  & 0.2040  & 0.4220  & 15.7437  & 0.0223  & 0.1493  & 0.7517  & 371.7175  \\
    Mamba & \textbf{0.0188} & \textbf{0.1371} & \textbf{0.2671} & \textbf{14.8583} & \textbf{0.0370} & \textbf{0.1924} & \textbf{0.2324} & \textbf{6.6232} & \textbf{0.0217} & \textbf{0.1472} & \textbf{0.4277} & \textbf{134.6655} \\
    \bottomrule
    \end{tabular}%
  \label{tab4}%
\end{table*}%

\begin{figure}[h]
\centering
\includegraphics[width=0.45\textwidth]{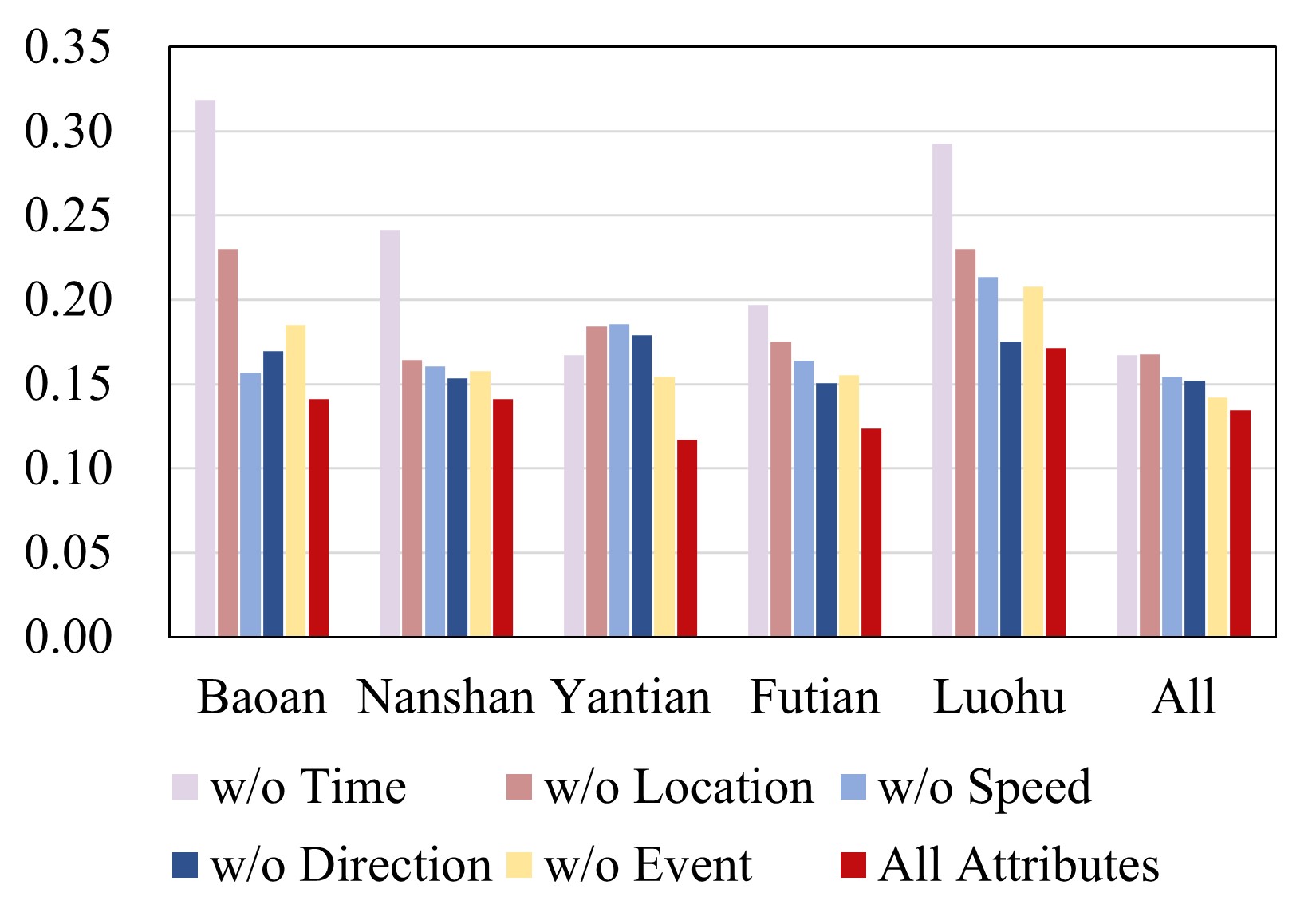}
\caption{Ablation analysis for AEM.}
\label{fig10}
\end{figure}

\begin{table*}[htbp]\scriptsize
  \centering
  \caption{Ablation analysis of SFM.}
         \renewcommand\arraystretch{0.8}
   \renewcommand\tabcolsep{4pt}
    \begin{tabular}{cccccccccc}
    \toprule
    \multirow{2}[4]{*}{\textbf{Methods}} & \multicolumn{3}{c}{\textbf{Baoan}} & \multicolumn{3}{c}{\textbf{Nanshan}} & \multicolumn{3}{c}{\textbf{Yantian}} \\
\cmidrule{2-10}          & \textbf{MSE} & \textbf{RMSE} & \textbf{MAPE} & \textbf{MSE} & \textbf{RMSE} & \textbf{MAPE} & \textbf{MSE} & \textbf{RMSE} & \textbf{MAPE} \\
    \midrule
    w/o FD & 0.0296  & 0.1721  & 0.6467  & 0.0270  & 0.1642  & 0.6541  & 0.0466  & 0.2158  & 2.1171  \\
    w/o $\mathcal{L}_S$ & 0.0243  & 0.1557  & 0.3954  & 0.0206  & 0.1436  & 0.9654  & 0.0160  & 0.1264  & 1.6188  \\
    \textbf{Ours} & \textbf{0.0199} & \textbf{0.1410} & \textbf{0.3509} & \textbf{0.0147} & \textbf{0.1212} & \textbf{0.4696} & \textbf{0.0136} & \textbf{0.1168} & \textbf{0.6826} \\
    \midrule
    \multirow{2}[4]{*}{\textbf{Methods}} & \multicolumn{3}{c}{\textbf{Futian}} & \multicolumn{3}{c}{\textbf{Luohu}} & \multicolumn{3}{c}{\textbf{All}} \\
\cmidrule{2-10}          & \textbf{MSE} & \textbf{RMSE} & \textbf{MAPE} & \textbf{MSE} & \textbf{RMSE} & \textbf{MAPE} & \textbf{MSE} & \textbf{RMSE} & \textbf{MAPE} \\
    \midrule
    w/o FD & 0.0251  & 0.1585  & 0.3945  & 0.0416  & 0.2040  & 0.4220  & 0.0226  & 0.1504  & 0.7598  \\
    w/o $\mathcal{L}_S$ & 0.0260  & 0.1614  & 0.2537  & 0.0344  & 0.1854  & 0.4117  & 0.0188  & 0.1372  & 0.4710  \\
    \textbf{Ours} & \textbf{0.0153} & \textbf{0.1236} & \textbf{0.2387} & \textbf{0.0294} & \textbf{0.1713} & \textbf{0.3573} & \textbf{0.0181} & \textbf{0.1345} & \textbf{0.4410} \\
    \bottomrule
    \end{tabular}%
  \label{tab2}%
\end{table*}%

\begin{figure*}[h]
\centering
\subfloat[MetaTTE.]{
\includegraphics[width=0.45\textwidth]{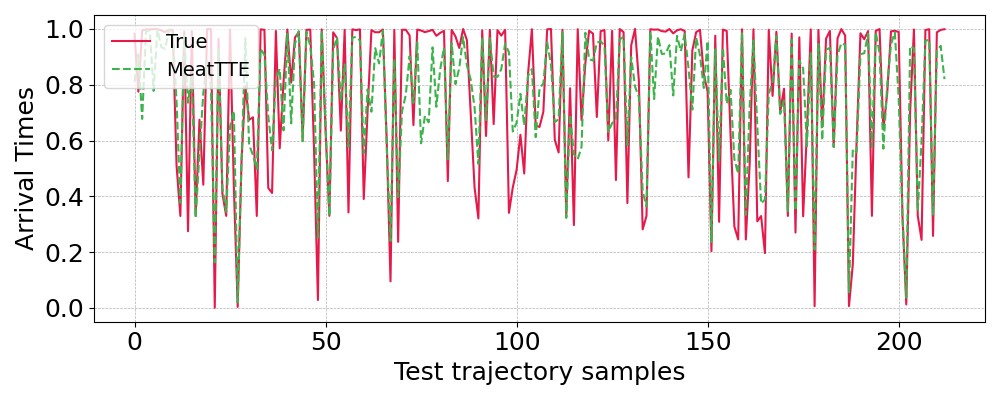}
}
\centering
\subfloat[IGT.]{
\includegraphics[width=0.45\textwidth]{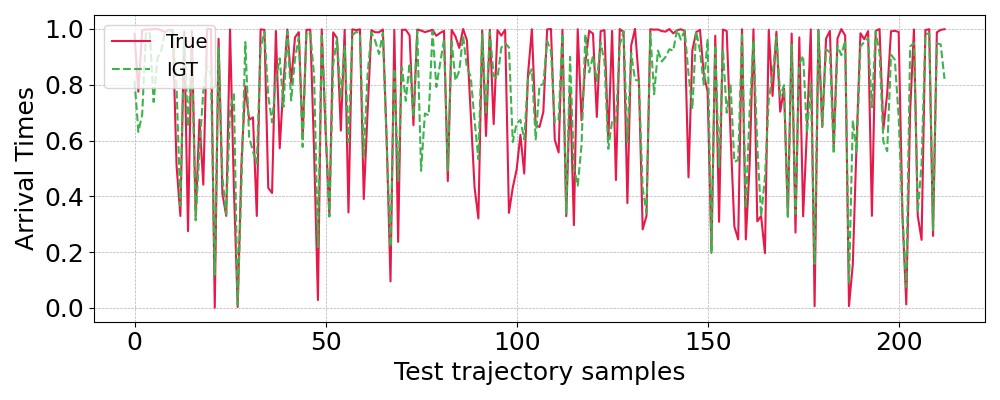}
}

\centering
\subfloat[Ours.]{
\includegraphics[width=0.45\textwidth]{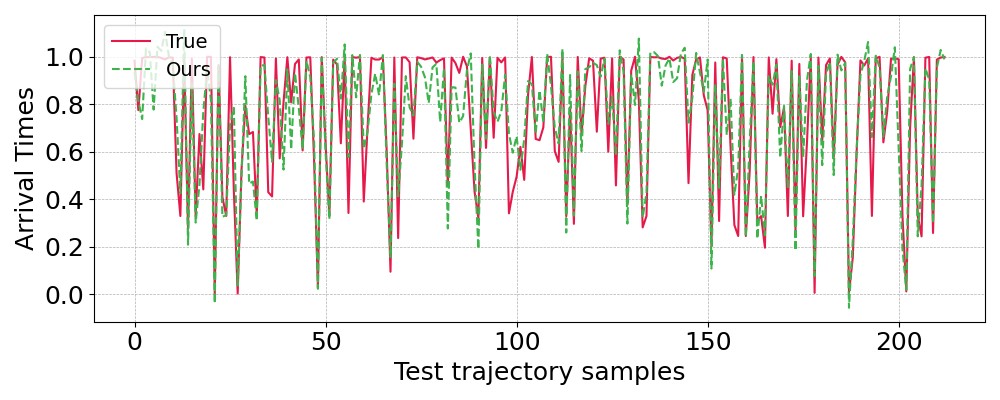}
}
\caption{ETA visualization for Nanshan dataset.}
\label{fig11}
\end{figure*}

\begin{figure*}[h]
\centering
\subfloat[$K$ and $l$ on Baoan dataset.]{
\includegraphics[width=0.25\textwidth]{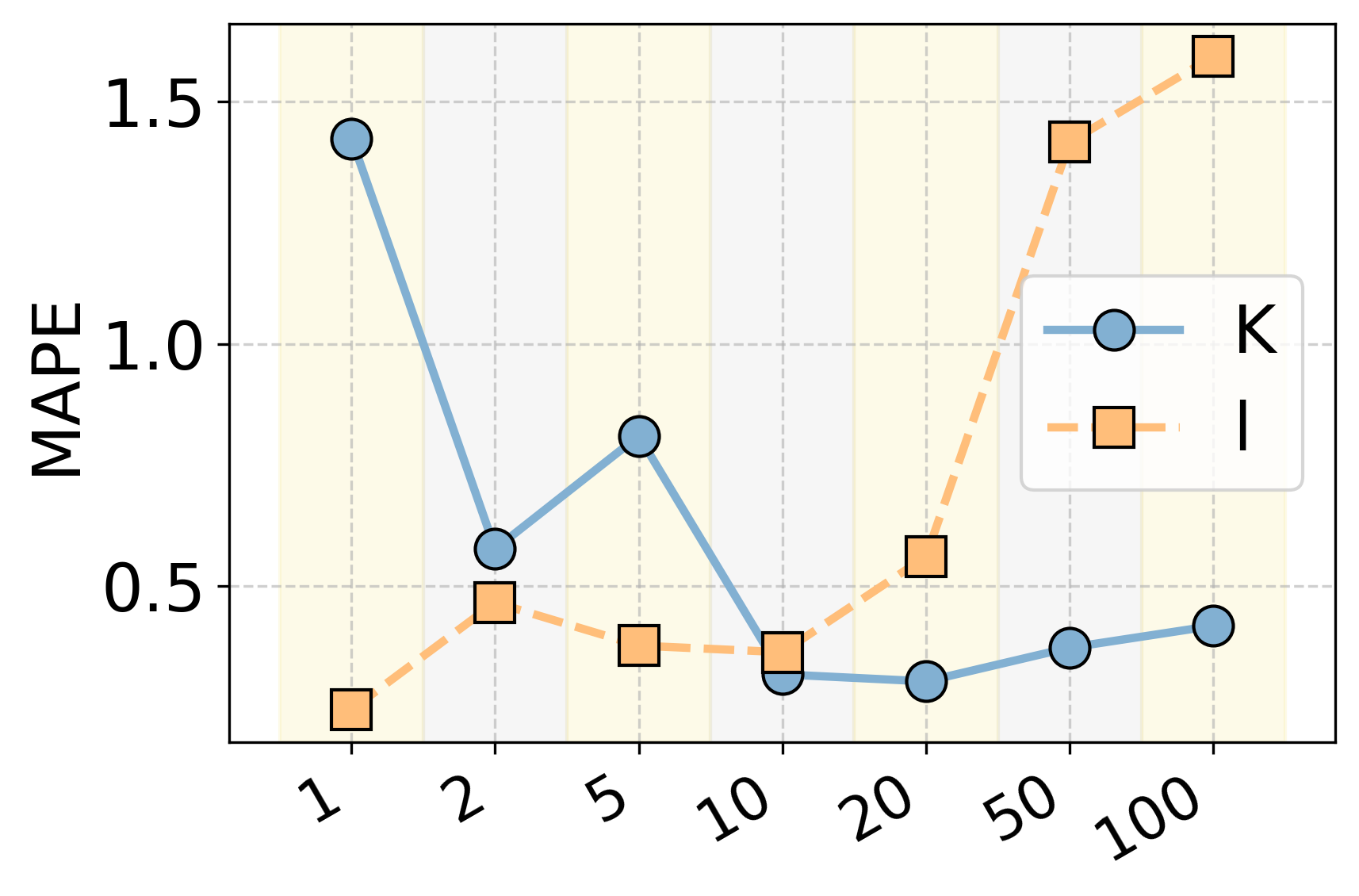}
}
\subfloat[$K$ and $l$ on Nanshan dataset.]{
\includegraphics[width=0.25\textwidth]{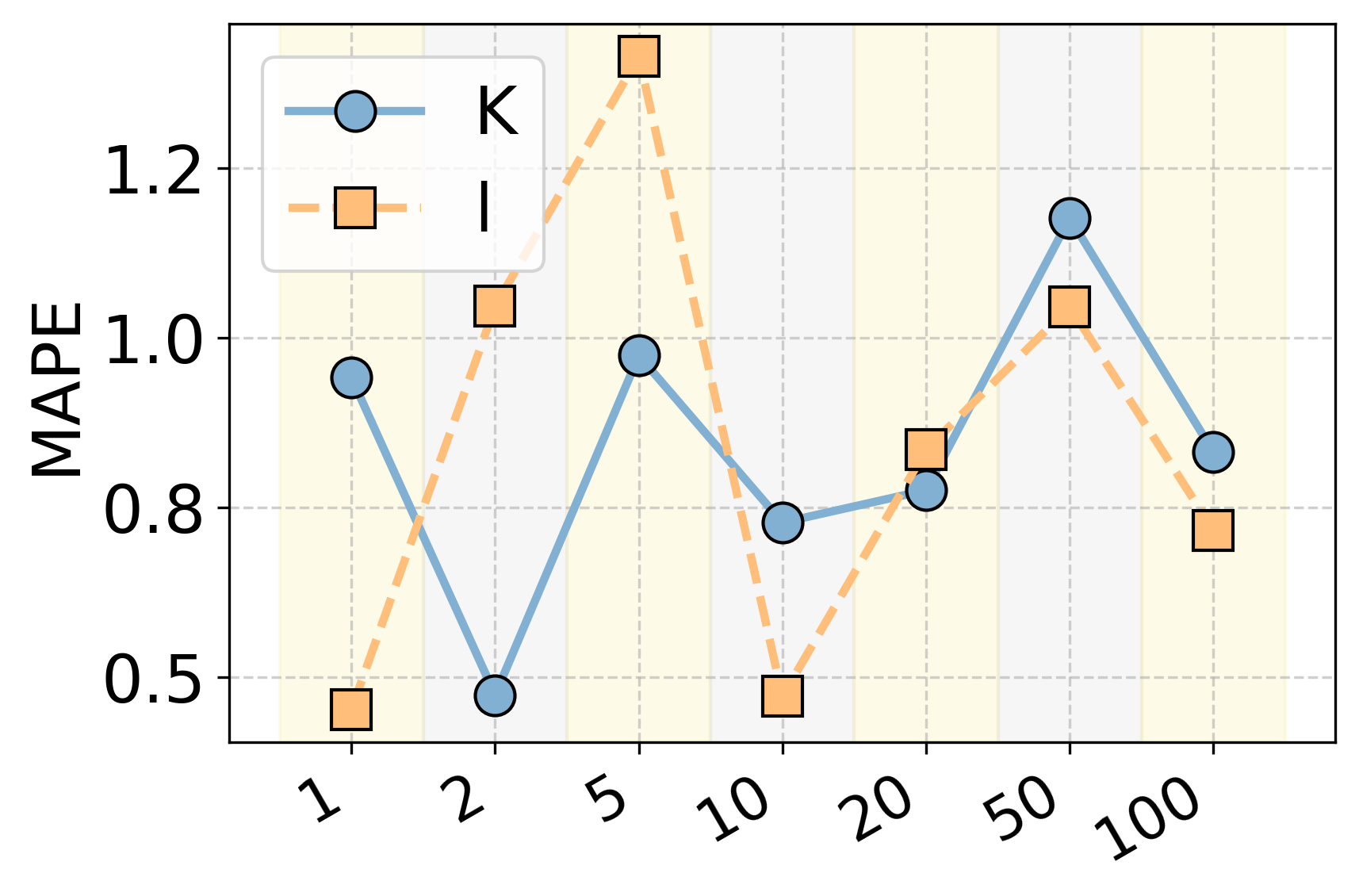}
}
\subfloat[$K$ and $l$ on All dataset.]{
\includegraphics[width=0.25\textwidth]{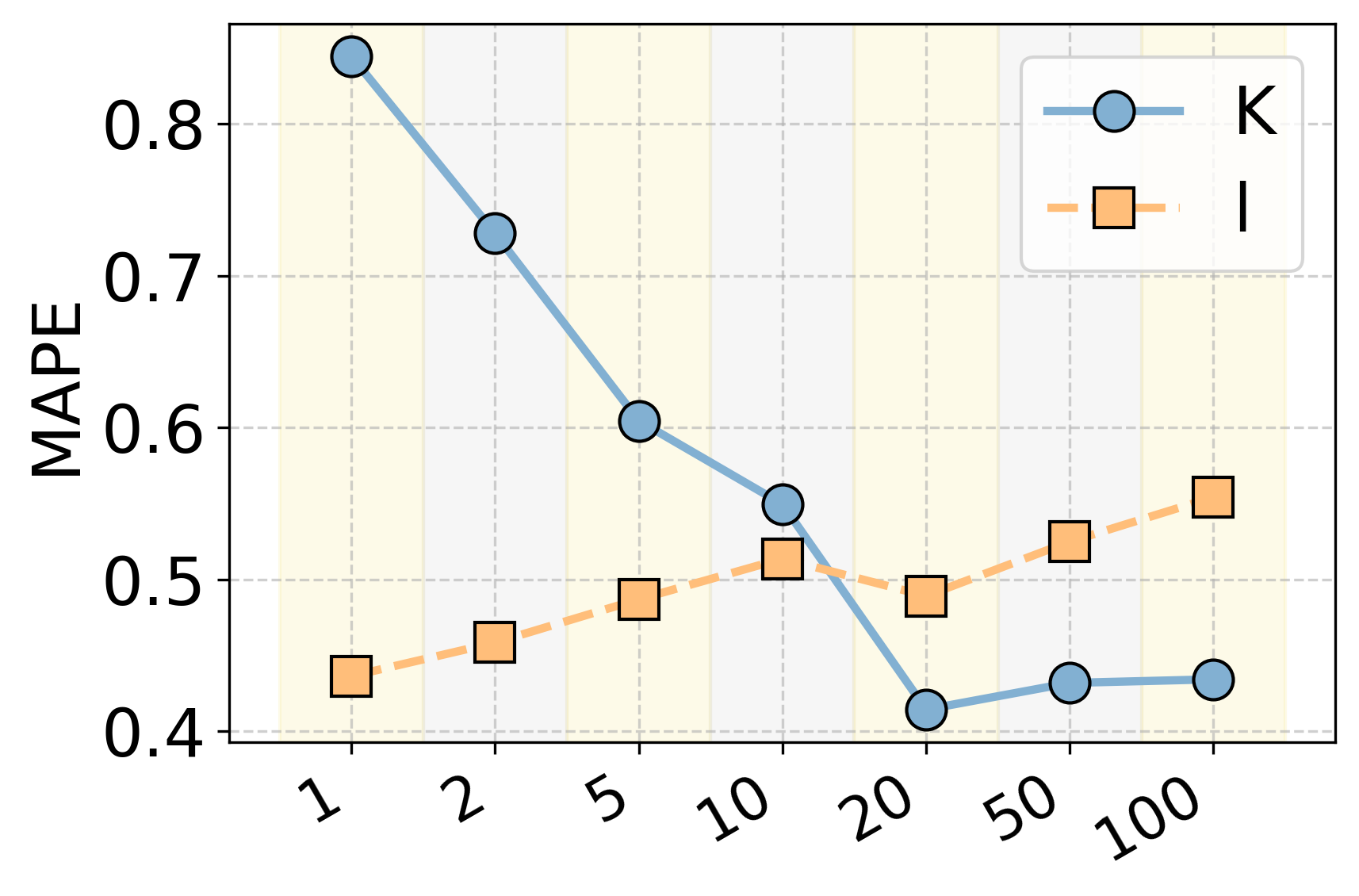}
}

\subfloat[$\eta$ and $\alpha$ on Baoan dataset.]{
\includegraphics[width=0.25\textwidth]{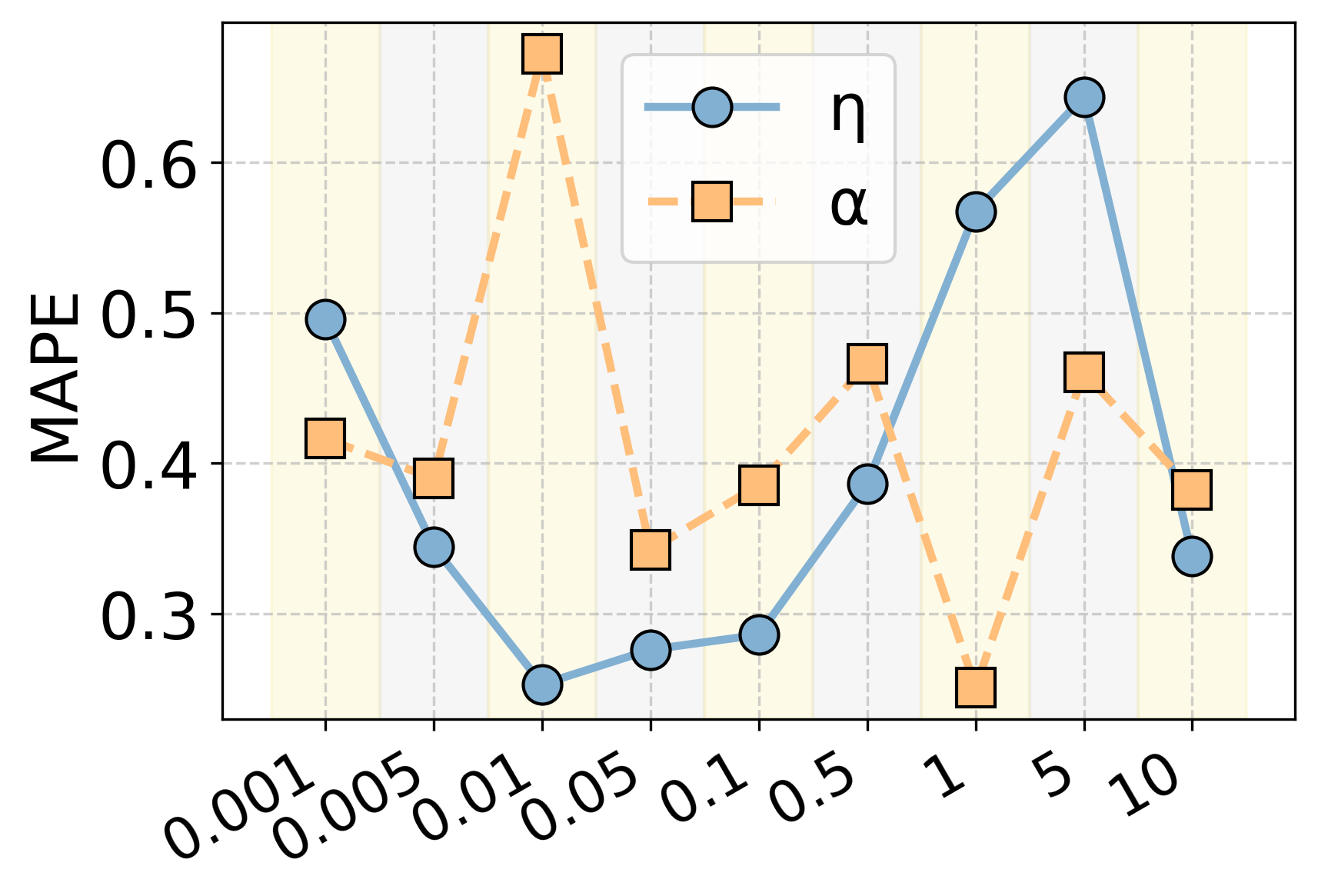}
}
\subfloat[$\eta$ and $\alpha$ on Nanshan dataset.]{
\includegraphics[width=0.25\textwidth]{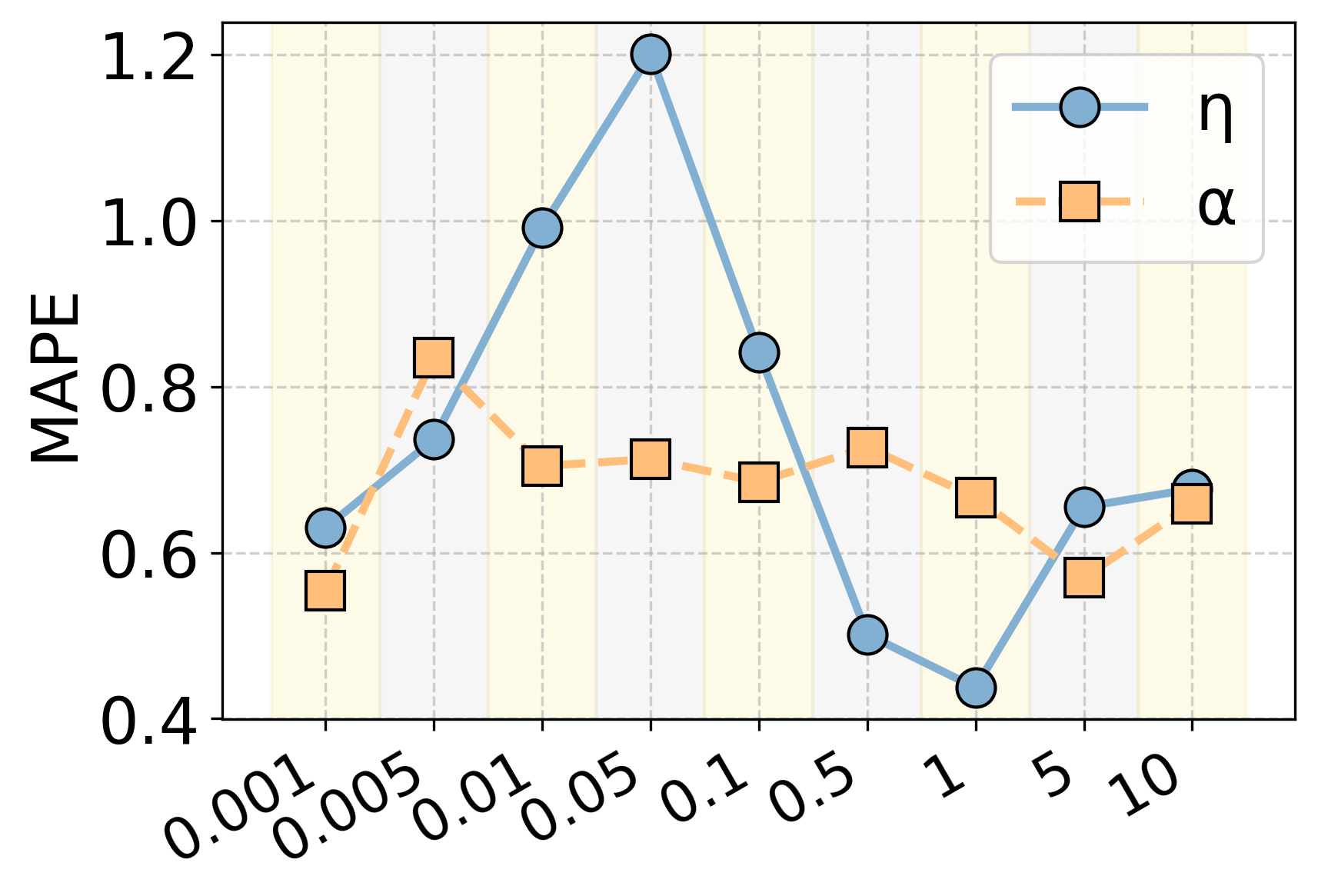}
}
\subfloat[$\eta$ and $\alpha$ on All dataset.]{
\includegraphics[width=0.25\textwidth]{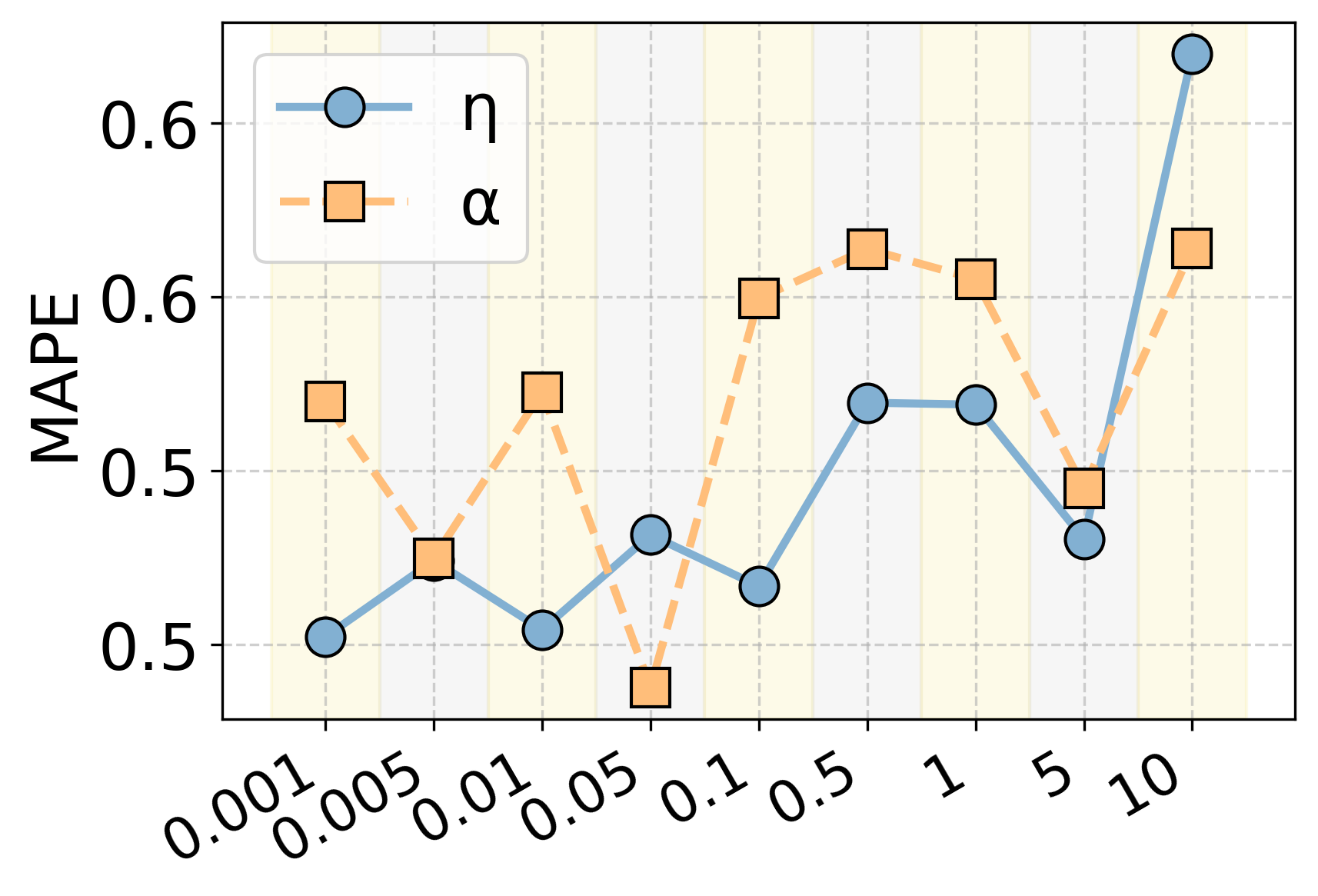}
}
\caption{{Hyperparametric experimental validation.}}
\label{fig13}
\end{figure*}

\subsection{Ablation Analysis}
Ablation studies are crucial in understanding the impact and significance of different components in a given model or methodology. In our research, we mainly focused on the TLM and AEM and SFM to evaluate their individual contributions.

\subsubsection{TLM Ablation Analysis}
The TLM leverages the state-of-the-art Mamba model, a selective state space model, for encoding complex time series data. To evaluate the unique contribution of Mamba, we conduct an ablation study by replacing it with alternative models such as RNN, LSTM, and Transformer, with results summarized in Table \ref{tab4}. The experiments reveal that Mamba significantly outperforms the other models across all evaluation metrics, demonstrating both superior accuracy and runtime efficiency. Particularly, Mamba's computational efficiency is evident, showing a marked advantage over Transformer, which incurs higher computational costs with longer sequences. Mamba’s ability to handle long-sequence temporal dependencies while maintaining computational efficiency highlights its value in improving ETA, making it a preferable choice for real-time, large-scale applications.

\subsubsection{AEM Ablation Analysis}
In the AEM, we explore the effect of different feature types—time, location, speed, direction, and event-related attributes—on prediction accuracy. Figure \ref{fig10} illustrates the impact of removing each feature type individually. Our findings show that omitting any single attribute reduces estimation accuracy, with location and spatial features showing the most significant impact. This result highlights the critical role of spatial context in ETA. The combination of all features provides the highest accuracy, suggesting that integrating diverse attributes allows the model to develop a more comprehensive understanding of the trajectory dynamics. This synergy among attributes enhances the model’s robustness and predictive power, underscoring the necessity of multi-feature integration in complex logistic applications.

\subsubsection{SFM Ablation Analysis}
For the SFM, we examine the role of two critical components: the Feature Diffusion (FD) module and the spatio-temporal loss function ($\mathcal{L}_S$). As shown in Table \ref{tab2}, the removal of either component results in a notable drop in performance, underscoring their importance. The FD module enables SFM to propagate features effectively across nodes with high similarity, enhancing the model’s ability to capture global spatial patterns and refine temporal correlations. This feature propagation mechanism, in conjunction with $\mathcal{L}_S$, allows SFM to capture subtle spatio-temporal dependencies, thereby contributing to the model's predictive accuracy in truck arrival time estimation.

\subsection{Visual Analysis of ETA}
To demonstrate the effectiveness of our framework, we compare it against two hybrid methods, MetaTTE and IGT, on the Nanshan district dataset, showcasing the visualization of the estimating time of arrival in testing, as shown in Figure \ref{fig11}. Specifically, the horizontal axis in the chart represents the sample index, while the vertical axis indicates the normalized arrival time. In the legend, the actual arrival times are depicted by a red solid line, whereas the estimated arrival times by our method are shown with a green dashed line. Observation of the chart clearly shows that our method surpasses the two comparative methods in terms of accuracy, closely aligning with the actual values. Furthermore, our framework effectively captures sudden and anomalous values in the data, accurately estimating the inherent patterns and rhythms of arrival times.

\subsection{Analysis of Hyperparameter Validation}

Hyperparameter tuning plays a vital role in optimizing the performance of our feature propagation model, as we aim to identify the best combination of hyperparameters that enhance predictive accuracy. In this section, we examine the impact of key hyperparameters on the feature propagation method, using initial default values of $K=20$, $l=10$, $\eta=0.01$, and $\alpha=0.1$. The experimental results are summarized in Figure \ref{fig13}.

The hyperparameter $K$, which defines the number of nearest neighbor nodes used during feature propagation, effectively controls the local neighborhood for feature diffusion. The parameter $l$, which represents the number of propagation iterations, influences the depth of feature propagation across the graph structure. We tested values of $K$ and $l$ across a range from 1 to 100. From the results, we observe that neither parameter has a straightforward linear relationship with performance. This indicates that increasing the neighborhood scope ($K$) and propagation depth ($l$) introduces additional useful context but may also add noise, thus necessitating dataset-specific tuning. The parameters $\eta$ and $\alpha$ represent balance factors that adjust the influence of sequence loss $\mathcal{L}_{SE}$ and residual connections. We explored values from 0.001 to 10. Our analysis indicates that, while optimal values differ across datasets, each dataset demonstrates a clear trend.

\section{Conclusion}\label{s5}

In this study, we developed and validated the Temporal-Attribute-Spatial Tri-space Coordination (TAS-TsC) framework for accurate truck arrival time estimation. The TAS-TsC framework integrates the state-of-the-art selective state space model, Mamba, with innovative techniques for attribute feature extraction and spatio-temporal feature propagation.

Comparative analyses with existing time-series prediction models confirm TAS-TsC’s superiority in both accuracy and efficiency. Notably, TAS-TsC’s domain transfer and generalization capabilities allow it to adapt effectively to new and diverse environments, making it versatile for real-world applications. Through extensive ablation studies, we assessed the contributions of each model component, revealing that the depth and scope of feature propagation, as well as optimized hyperparameter configurations, play crucial roles in achieving robust ETA. 

Looking forward, we plan to explore real-time deployment of TAS-TsC in dynamic traffic conditions, incorporating live GPS data to evaluate performance in real-time ETA updates. We will also focus on refining the spatio-temporal relation graph to enhance its adaptability in varying traffic and geographic patterns. Additionally, extending TAS-TsC to multi-modal logistics, such as combining truck and rail ETA, may further expand its applicability and improve end-to-end logistics efficiency.

\bibliographystyle{elsarticle-num} 
\bibliography{asc}

\end{document}